This article has been accepted for publication in IEEE Access. This is the author's version which has not been fully edited and content may change prior to final publication. Citation information: DOI 10.1109/ACCESS.2024.3523325

# Towards Semi-Autonomous Robotic Arm Manipulation Operator Intention Detection from Force Data

**Abdullah S Alharthi [1], Ozan Tokatli [2], Erwin Lopez [3], and Guido Herrmann [3], Senior Member, IEEE**

[1]Electrical Engineering Department, College of engineering, King Khalid University, Abha, Saudi Arabia
[2]Remote Applications in Challenging Environments United Kingdom Atomic Energy Authority
[3]Electrical Engineering Department, The University of Manchester

Corresponding author: Abdullah S Alharthi (e-mail: Asalharthi@kku.edu.sa)

The authors extend their appreciation to the Deanship of Research and Graduate Studies at King Khalid University for funding this work through Large Research Project under grant number RGP2/254/45.

**ABSTRACT** In hazardous environments like nuclear facilities, robotic systems are essential for executing tasks that would otherwise expose humans to dangerous radiation levels, which pose severe health risks and can be fatal. However, many operations in the nuclear environment require teleoperating robots, resulting in a significant cognitive load on operators as well as physical strain over extended periods of time. To address this challenge, we propose enhancing the teleoperation system with an assistive model capable of predicting operator intentions and dynamically adapting to their needs. The machine learning model processes robotic arm force data, analyzing spatiotemporal patterns to accurately detect the ongoing task before its completion. To support this approach, we collected a diverse dataset from teleoperation experiments involving glovebox tasks in nuclear applications. This dataset encompasses heterogeneous spatiotemporal data captured from the teleoperation system. We employ a hybrid Convolutional Neural Network (CNN) and Long Short-Term Memory (LSTM) model to learn and forecast operator intentions based on the spatiotemporal data. By accurately predicting these intentions, the robot can execute tasks more efficiently and effectively, requiring minimal input from the operator. Our experiments validated the model using the dataset, focusing on tasks such as radiation surveys and object grasping. The proposed approach demonstrated an F1-score of 89% for task classification and an F1-score of 86% classification forecasted operator intentions over a 5-second window. These results highlight the potential of our method to improve the safety, precision, and efficiency of robotic operations in hazardous environments, thereby significantly reducing human radiation exposure.

## I. INTRODUCTION

Nuclear decommissioning is a complex, long-term process that involves the careful dismantling of a nuclear facility and the safe disposal of radioactive materials. This process ensures that nuclear sites are properly cleaned up and decontaminated, protecting both the environment and human health from the dangers posed by residual radioactive substances. In the UK, the nuclear decommissioning program is the largest environmental remediation effort in the world. It is expected to take over 100 years and costs billions of pounds to complete [1]. The extensive duration and scale of this project highlights the significant challenges involved in safely managing nuclear materials over such a long period. One key aspect of nuclear decommissioning is the use of gloveboxes, which are sealed containers that allow professional operators to handle and dismantle radioactive materials without direct contact. These gloveboxes are designed to protect workers from exposure to radiation while still enabling precise manipulation of hazardous substances. However, over time, gloveboxes can become contaminated due to the radioactive materials they contain. This contamination increases the risk of radiation exposure to operators, especially given the prolonged nature of the decommissioning process [2].

To address the risks and improve both safety and efficiency in nuclear decommissioning, industry has increasingly adopted robotic systems and advanced technologies. These innovations enable the remote handling of nuclear waste, reducing the need for human operators to perform dangerous tasks directly. Teleoperated robotic systems have been highly effective in managing radioactive materials inside gloveboxes. By allowing operators to control the robots from a safe distance, these systems greatly reduce the risk of radiation exposure and prevent the spread of contamination outside the glovebox [3], [4]. Figure 1a shows how the proposed teleoperated robots are used within a glovebox during the decommissioning process, making the procedure safer and more efficient [4].

### A. MOTIVATION

The current robotic arm used inside the glovebox [4] relies heavily on the operator's ability to control it via a haptic device (as shown in Figure 1b), guiding the robot's end effector through a series of tasks. The operator must manually convey their intent by precisely directing the robotic arm through each step of the manipulation process. This approach demands a high level of concentration, coordination, and spatial reasoning, making the process both labor-intensive and mentally taxing. Over time, the need for constant human input and precise control can lead to operator fatigue, reduced efficiency, and an increased risk of errors. The cognitive burden placed on the operator to manage task execution while ensuring accurate intent inference can create bottlenecks, particularly in environments requiring real-time decision-making and precision [5]. This dependence on manual control also limits the potential for seamless human-robot collaboration, as the effort required reduces the system's agility and responsiveness in adjusting the robotic arm's position for subsequent tasks. As a result, this approach may fall short in environments demanding swift and precise task execution, such as hazardous glovebox operations, where real-time adjustments and continuous monitoring are critical.

To address this issue, recent studies have shown that humans can recognize others' intentions from observations, demonstrating that intentions can be inferred from non-verbal communication [6]. Therefore, recognizing the human operator's intentions can aid in augmenting the human control of the robot in a shared control operation. To facilitate faster and more natural interaction for a sequence of tasks, non-verbal cues and indirect signals that the user implicitly provides when performing tasks using robot controllers can be recognized using machine learning to predict specific scenarios. The recognition of these scenarios can be dedicated to understanding, designing, and evaluating robotic systems for use by or with humans [7].

In a glovebox, a haptic force-feedback manipulator, a haptic device (Virtuose 6D TAO) [26], moves the robotic arm in the glovebox to mimic the haptic device movement with force data in all directions. The robotic arm provides force feedback to the haptic device to express the robotic arm's state to help the user recognize the limitation of the arm movement. However, the continuous adjustments required to control the haptic device can lead to variability in the operator's behavior, affecting their ability to consistently maneuver the robotic arm's end effector to complete tasks accurately. In such cases, the operator must disconnect the haptic device interface with the robotic arm to correct the position of the haptic device before reconnecting to the robotic arm, further interrupting the workflow and increasing task complexity. Consequently, deploying traditional and proven methods, such as recording the operator motion using IMU sensors, motion capturing sensors, or using haptic device information, does not benefit the detection of the operator's intention.

### B. OBJECTIVES

In this work, we propose a machine learning approach to detect the operator's intention in robot-assisted glovebox telemanipulation using only the robotic arm force sensing towards semi-autonomous manipulation. An open-source dataset featuring autonomous manipulation of various objects was obtained to test the resilience of robotic arm force data in detecting operator intentions [8]. The robust results encouraged the recording of the robotic arm force data as spatiotemporal data, in which the state of each joint is recorded in time frames. Given this recorded spatiotemporal information, a machine learning model is built using CNN and Long Short-Term Memory (LSTM) to process raw spatiotemporal data from a robotic arm, aiming to accurately detect the operator’s intentions. The model relies solely on the robotic arm's force feedback, eliminating variability caused by environmental conditions or operator inconsistencies. By focusing on joint positions, velocities, and forces, the model forecast spatiotemporal data to predict the operator’s intentions while in task execution. In such the operator's intentions are derived from the model prediction based on forecasted data, such as radiation survey and object grasping.

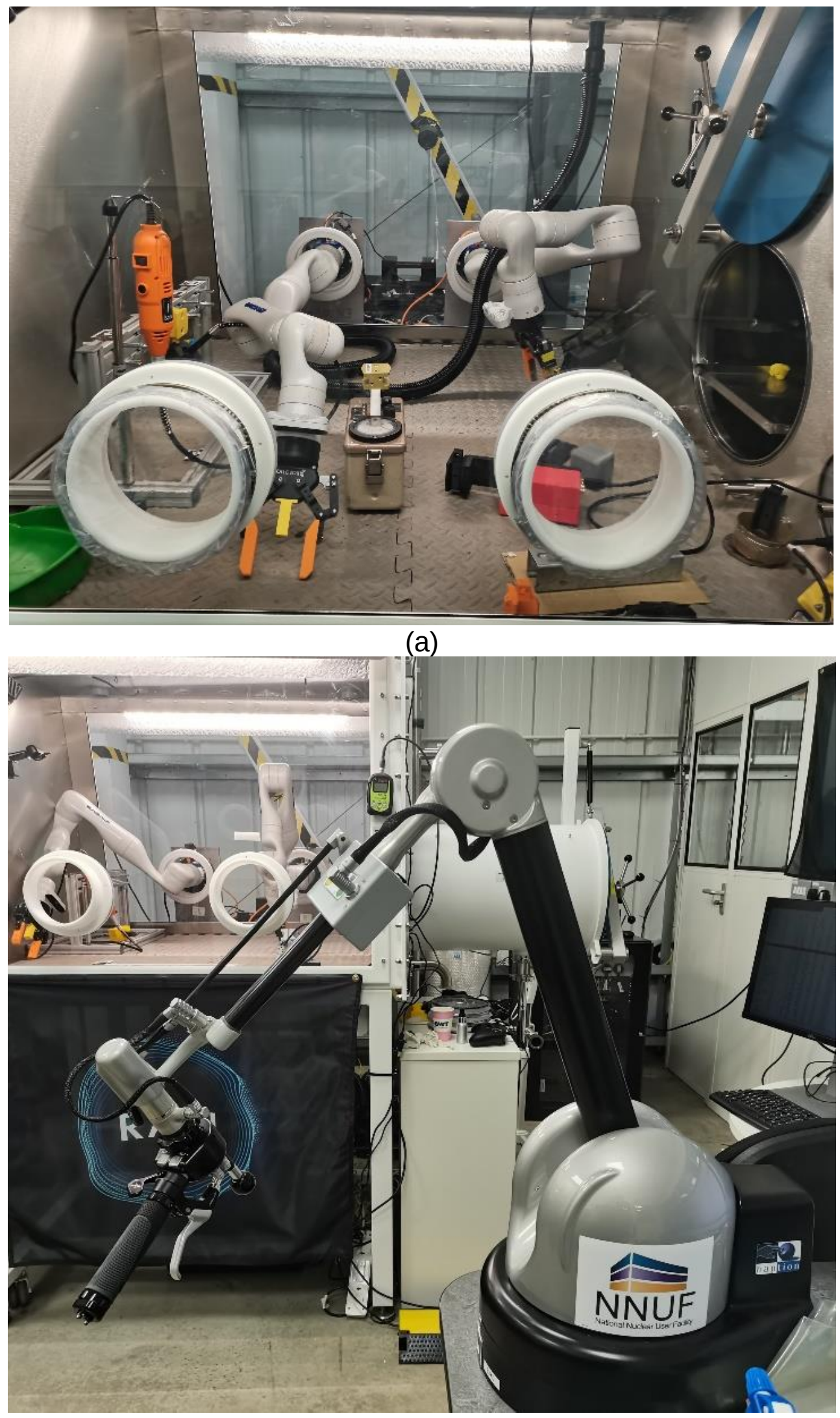

(a)

(b)

**FIGURE 1. Tele-manipulation systems in glovebox environment. (a) two Kinova robotic arms in the Glovebox environment, (b) haptic device (Virtuose 6D TAO) to control robotic arms.**

### *C. KEY CONTRIBUTIONS*

Our approach aims to study the effectiveness of operator intention detection using only the robotic arm force sensing towards semi-autonomous manipulation. This approach would potentially limit the control burden in teleoperated manipulation during complex glovebox operations, such as radiation survey and object grasping, by detecting what the operator wants to do and later executing it autonomously or activate the other robotic arm in the glovebox (see Figure 1 a) to assists autonomously. The results of this study have the potential to significantly improve the safety and efficiency of nuclear decommissioning processes, reducing the risk of human exposure to radiation while also improving the accuracy and speed of robotic operations.

The main contributions of this research are summarized as follows:

- *Feature Extraction from Robotic Arm Data*: Developed advanced tools to extract crucial features from the robotic arm’s spatiotemporal data—such as joint positions, velocities, and force measurements—without relying on external sensors like cameras or IMUs. This approach enhances data reliability and analysis accuracy, especially in complex, challenging environments where external sensing may be impractical or limited.
- *CNN-Based Data Transformation for Enhanced Forecasting*: Utilized a CNN architecture to process spatiotemporal signals from the robotic arm, transforming raw data into structured, meaningful information suitable for forecasting. This method preserves essential temporal patterns while reducing noise, enabling the system to make more informed predictions, ultimately addressing the need for improved decision-making capabilities in uncertain or harsh conditions.
- *Task Prediction for Improved Robotic Arm Adaptability*: By leveraging extracted temporal data for both classification and prediction, the model effectively forecasts remaining task sequences based on partial observations. For example, upon processing the initial segment of a task, the system accurately predicts subsequent actions required to complete it, thus enabling the robotic arm to adapt and respond proactively to task demands, contributing to its reliable operation in challenging and unpredictable scenarios.

The following section II discusses related work in literature and section III describes the technical background of the recorded data and CNN model. Our specifically implemented networks and their classification performance are presented in section IV. Finally, our results are discussed in section V, followed by the conclusion in Section VI.

## II. RELATED WORK

Understanding intention inference, which involves grasping the intended goal, target, action, or behavior, is crucial for achieving smooth interactions with robots. This has become a focal point within research communities, particularly in the expansive field of robotics where there is a growing emphasis on developing safe integration methods for robots functioning as companions and collaborators in diverse work settings [9].

Researchers have consistently explored innovative approaches across various areas. For instance, there is active exploration in the development of semi-autonomous methods for robotic arm manipulation. These methods, utilizing technologies such as vision and force sensing, facilitate telemanipulation with primary-replica tele-robot systems [10], offering protection for workers in hazardous environments.

Beyond advanced sensing and telemanipulation, researchers have delved into human activity recognition through computer vision [11] to predict human intention in controlling robotic arms. Surgical gestures can be segmented and recognized using Hidden Markov Models, enabling robots to comprehend and respond to human actions in medical scenarios [12]. This breakthrough enhances collaboration between surgeons and robotic assistants, leading to improved surgical outcomes and patient safety.

Gaze patterns have also been extensively studied to extract valuable information for human-robot shared manipulation. Models predicting human intentions have been developed by analyzing features like saccades, fixations, smooth pursuits, and scan paths, guiding robot behaviors accordingly [13]. This understanding enables robots to anticipate human actions, resulting in smoother and more efficient collaborative tasks, with applications ranging from industrial manufacturing to healthcare and rehabilitation.

Building on these advancements, anticipatory control methods have been explored, allowing robots to proactively perform tasks based on predicted human actions observed through gaze patterns [14]. In scenarios like human-robot assembly lines, this proactive behavior minimizes delays and optimizes workflow by predicting the next action of the human worker.

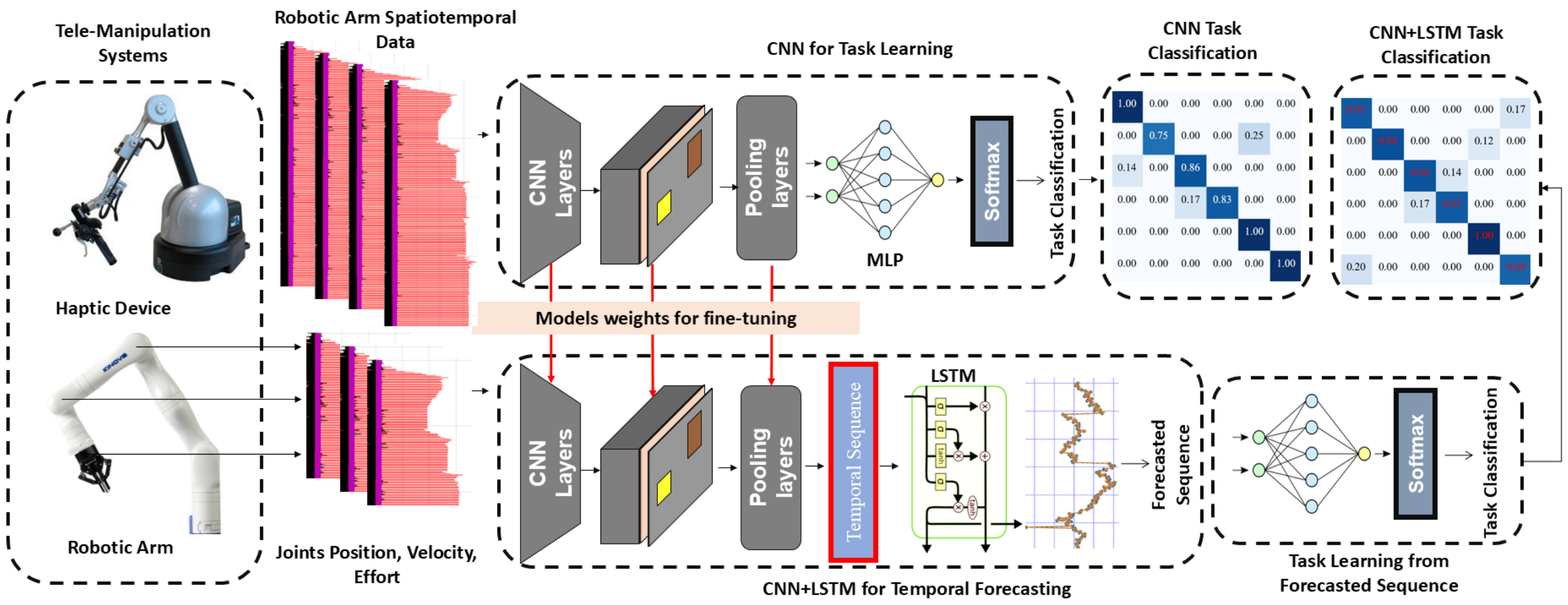

**FIGURE 2. Proposed framework for robotic arm spatiotemporal data forecasting and classification.**

Moreover, researchers have investigated incorporating gestures and language expressions to enhance human-robot communication. Systems utilizing multimodal Bayesian filters interpret a person's referential expressions to objects, enabling robots to understand and respond more effectively to human commands [15]. This development facilitates intuitive interactions between humans and robots, breaking down communication barriers and enabling robots to seamlessly integrate into various social and domestic settings.

For teleoperation scenarios, where human operators and robotic systems converge, there lies a promising avenue for the amalgamation of human cognitive prowess with the precision and dependability inherent in robotic technology. This convergence, often realized through shared control mechanisms, represents a symbiotic relationship where both human intellect and machine efficiency synergize to achieve common goals [16], [17]. Recent explorations by various researchers [18], [19] have delved deep into the realm of shared control, employing optimization techniques to discern and fulfill user objectives effectively. These endeavors aim not only to assist users but also to seamlessly integrate human intentions into the control paradigm. While existing teleoperation architectures, such as those described in [20], have made strides in leveraging learned trajectory properties to aid in reaching tasks, they have yet to fully incorporate the nuanced intent of the operator into their frameworks. However, emerging insights, as exemplified in [21], suggest that intent recognition holds the key to tailoring shared control autonomy to the specific needs and preferences of the human operator. This paradigm shift reframes the selection of control autonomy as a nuanced role distribution problem [22], [23], wherein the balance between human and robot autonomy is delicately calibrated to optimize task performance while ensuring user comfort and efficiency. Furthermore, the recent framework proposed in [24] introduces a novel approach, harnessing the power of Gaussian mixture models and Probabilistic Movement Primitives to infer optimal trajectories for reaching specific targets. This innovative methodology not only showcases the potential for enhancing task execution but also underscores the importance of integrating advanced modeling techniques into shared control architectures.

However, challenges arise when considering contaminated areas with radiation, particularly in glovebox environments. Prolonged operator exposure can lead to hazardous situations, rendering the application of computer vision and gestures impractical in such scenarios.

Addressing these challenges, intention inference within the context of glovebox telemanipulation offers a novel approach to enhance the safety and efficiency of nuclear decommissioning processes. Unlike traditional telemanipulation methods, where haptic devices control robotic arms in harsh environments, gloveboxes, characterized by their highly unstructured nature and significant uncertainty, require a specialized approach. The integration of operator intention becomes crucial in guiding robotic actions and achieving seamless human-robot.

## III. MATERIALS AND METHODS

In this study intention detection is handled as supervised learning, where data recording and manual labeling are employed to train a neural network to predict and forecast the operator's intentions. The primary objective is to develop a model capable of recognizing the operator's intentions based on task observations. Data analysis and experiments are conducted to validate this intention recognition methodology, with the potential for implementing real-time intention recognition during telemanipulation. Figure 2 illustrates the proposed framework for task forecasting and classification. The following sections detail the framework, including data recording protocols, data processing, and the CNN, LSTM, and hybrid CNN+LSTM model architectures.

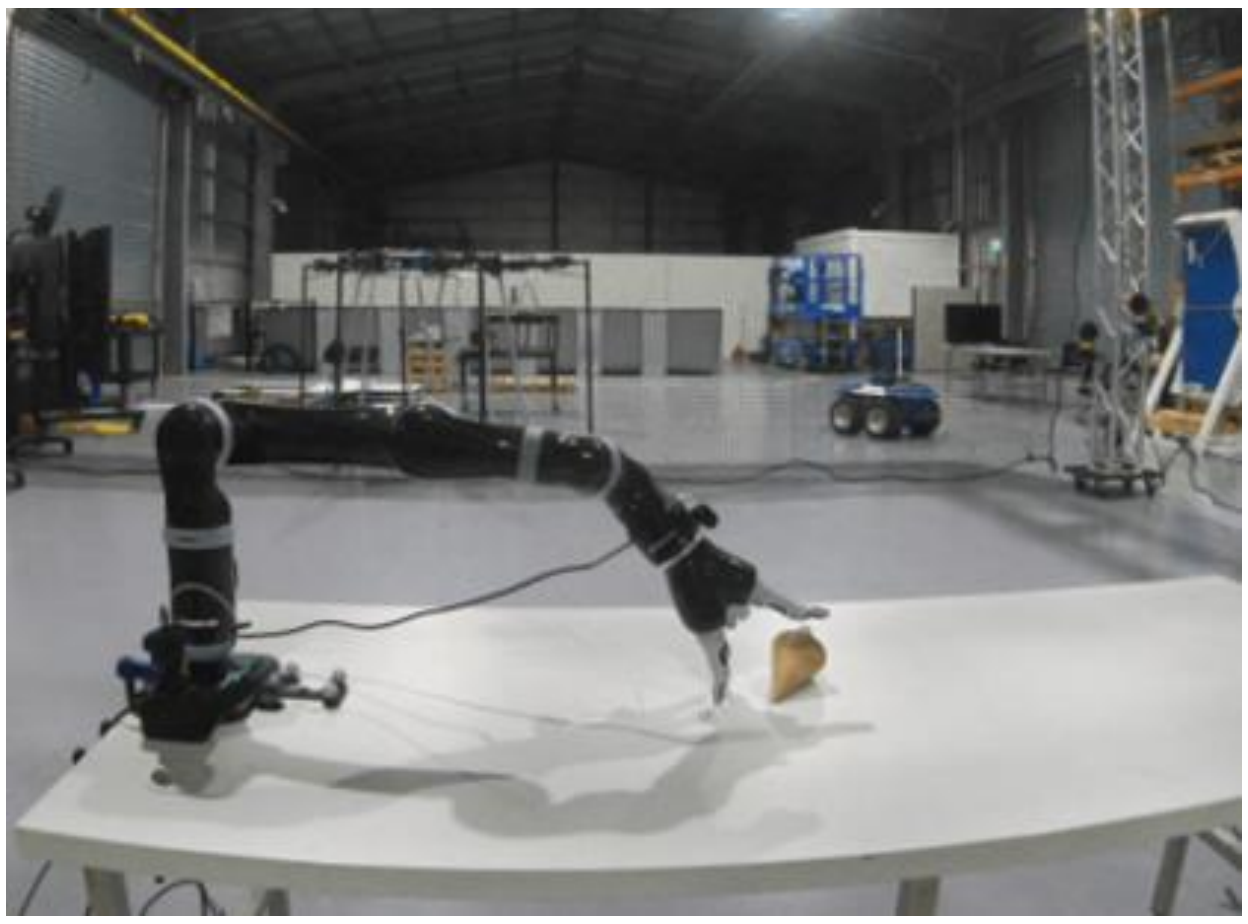

**FIGURE 3. Kinova-mic robotic arm in dataset 1 with fully autonomous manipulation of objects [8].**

### *D. Data*

In this study two sets of data were used to validate the intentions recognition methodology. Initially an open-source dataset is acquired and processed followed by data recording from the glovebox robotic arm as detailed in the following.

1) ***Dataset 1***

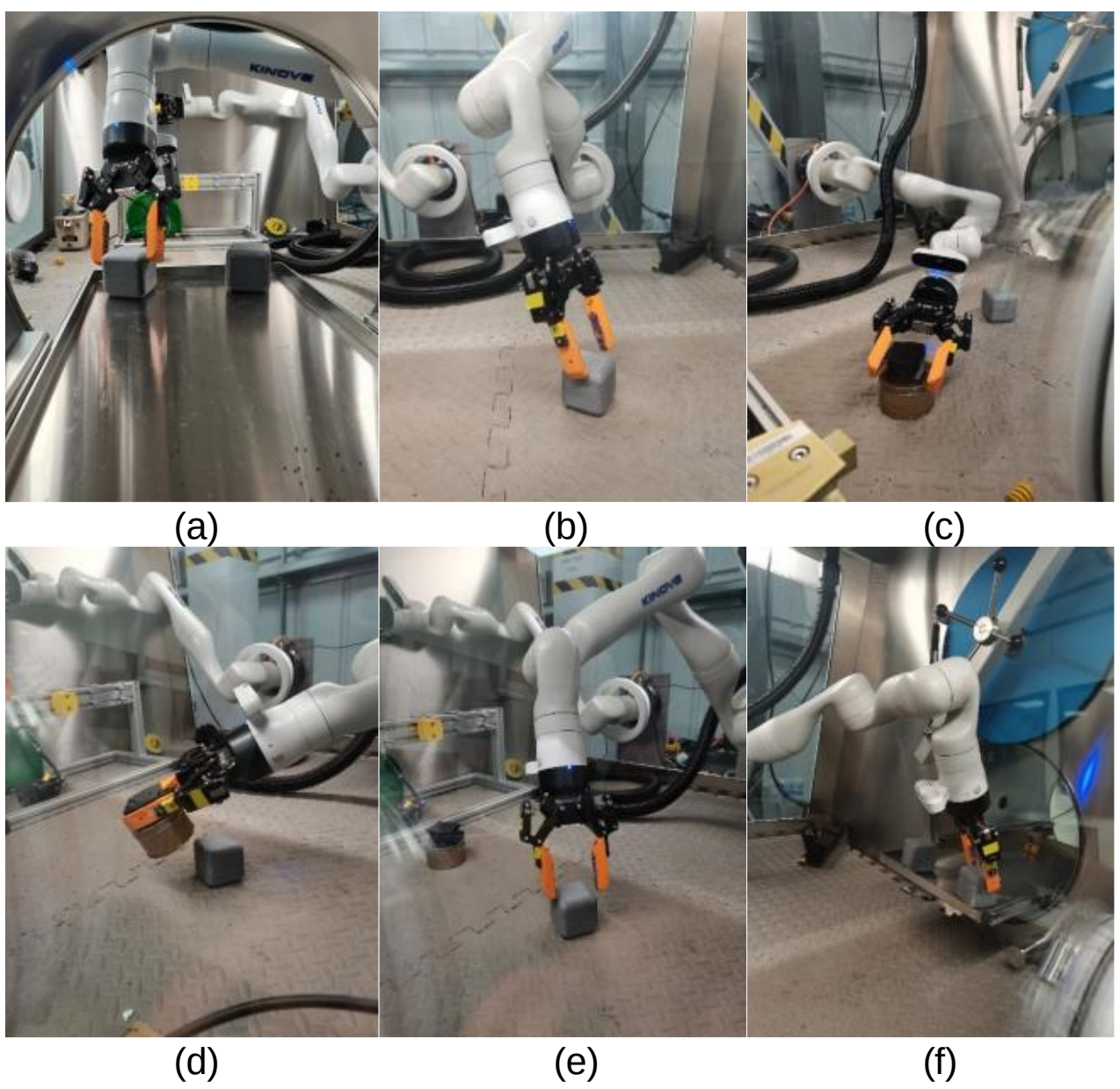

**FIGURE 4. Object posting in glovebox and radiation surveying for 6 tasks. From left to right Post object into the glovebox, Place object on the glovebox floor, Grasp the radiation sensor, Radiation survey, Return the radiation sensors, Grasp the object, and post it out of the glovebox.**

TABLE I
OPEN-SOURCE DATA CLASSES

| Task | Description | Number of samples |
|---|---|---|
| 1 | Kinematic motion of the arm, Simple | 20 |
| 2 | Kinematic motion of the arm, Complex | 20 |
| 3 | Plastic & Wooden cube pushing | 40 |
| 4 | Plastic & Wooden cylinder pushing | 40 |
| 5 | Plastic & Wooden cone pushing | 40 |
| 6 | Plastic & Wooden cuboid pushing | 40 |

An open-source benchmark of autonomous manipulation (automatic motion of the robotic arm) of objects with one robotic arm [6] is acquired to test the CNN ability to recognize the manipulation tasks of objects. The benchmark is recorded using a series of joint positions and rotations with corresponding times protocols to complete each task autonomously [6]. The tasks as detailed in Table 1 comprise of 210 samples, 20 samples for each task, recorded from kinova mico arm (see Figure 3) for 10 seconds:

1. 6 degrees of freedom of the robotic arm pose (translation and rotation: translation in x,y,z and rotation as a quaternion). Recorded at 100 Hz.
2. Force torque data (Robotiq FT300 force torque sensor mounted at the wrist) recorded at 100Hz, force in the x,y and z axis along with moments in the x,y and z axis).
3. Joint Torques (recorded at 100 Hz for the 6 joints of the Kinova).
4. Finger positions (three fingers) recorded at 100 Hz, for the three fingers.

The final dataset is the mean data for the arm pose, force torque sensor, joint torques and finger positions. In this data the labelling processed is based on categorizing the data manually based on each sample manipulation motion. In other words, Robot Operation System (ROS) bag file is labelled upon the task completion, which is converted to CSV file to create the benchmark, and the label is the file name with trial number.

2) ***Dataset 2***

Object manipulation and radiation survey with bilateral telemanipulation in glovebox. The data is recorded with one operator for object posting in and out of the glovebox and radiation survey of the object using one Kinova arm. The data are recorded for 6 tasks (see Figure 4), 20 samples for each task. The tasks are titled as follows.

1. Post object into the glovebox.
2. Place object on the glovebox floor.
3. Grasp the radiation sensor.
4. Radiation survey.
5. Return the radiation sensor.
6. Grasp the object and post it out of the glovebox.

Dataset 2 tasks are recorded in succession, e.g. task 1: start recording "extend the robotic arm to grasp object from the glovebox port and grasp the object" end the recording save the generate ROS bag file and label the file as task (1) trial (1); task 2: start recording "extend the robotic arm to place object on the glovebox floor" end the recording save the generate ROS bag file and label the file as task (2) trial (1). The tasks are repeated for number of 20 trials. The final data are converted to CSV, and it consists of one base and 7 joints position, velocity, effort, or forces each joint movement: a total of 24 spatial data points. The data are recorded at 100 Hz. The time duration for the tasks varies depending on the robotic arm previous position.

| | | | |
|---|---|---|---|
| 1 | 2 | 3 | 4 |
| 5 | 6 | 7 | 8 |
| 9 | 10 | 11 | 12 |

**FIGURE 5.** **Grid layout for a radiation surveying test.**

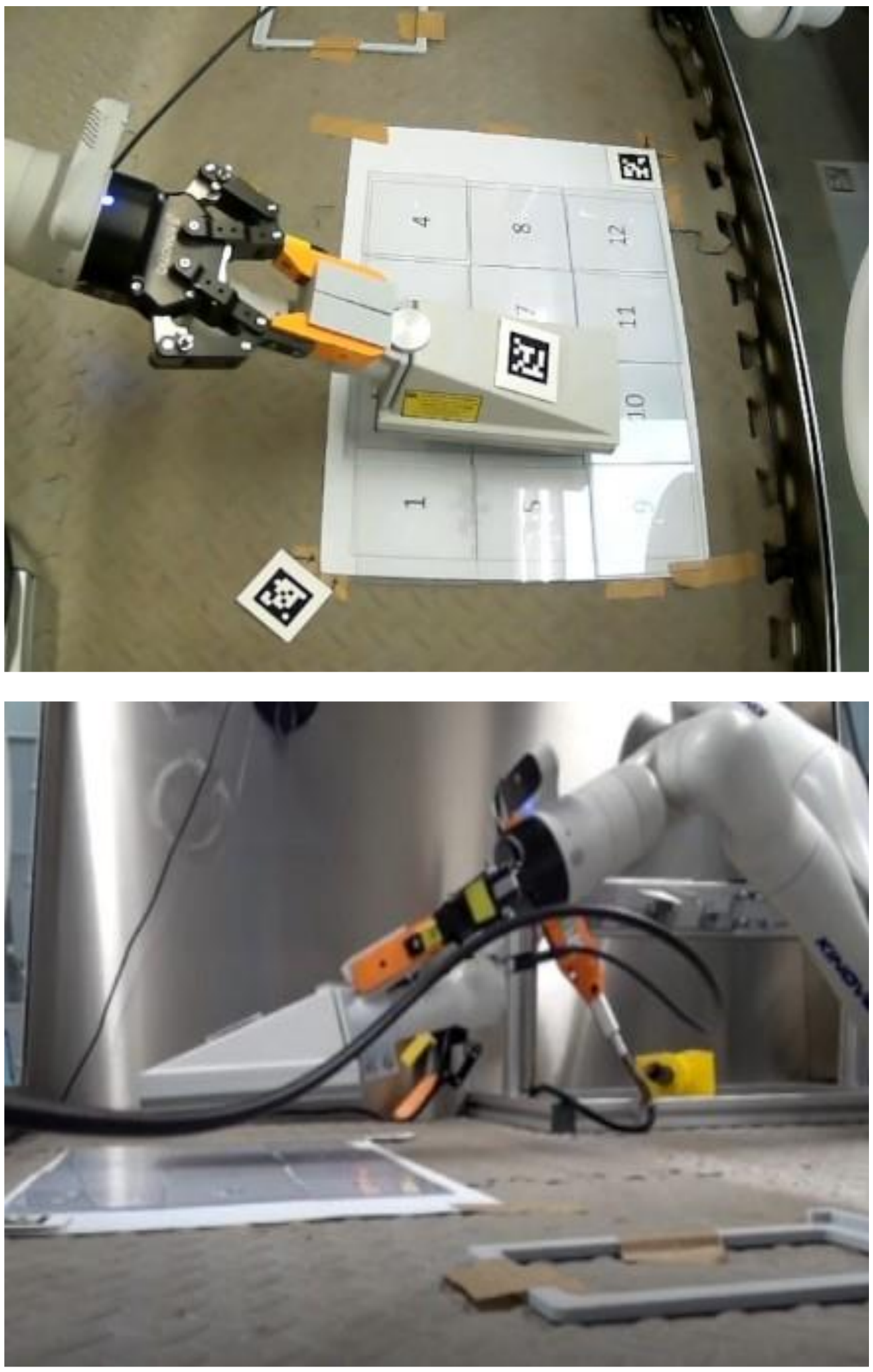

**FIGURE 6.** **Grid radiation survey of data (3).**

***3) Dataset 3***

This data consists of grid radiation survey (see Figure 5) using the tele-manipulation systems in the glovebox environment. The data was recorded for 6 operators using the tele-manipulation system consisting of a Kinova Gen3 manipulator and Haption Virtuose 6D haptic interface. 4 samples recorded for each operator while performing radiation survey see Figure 6, with a data comprise of 24 samples. A spatiotemporal signal is recorded as joints position, velocity, effort, a total of 32 spatial data point. The sample are recorded at 100Hz. Similar to the data recording protocol in dataset 2, the data recording starts when the survey a grid and ends when all the area is scanned. The duration of the radiation is varying and depends on the operator.

### *E. CNN model*

CNN is a special type of hierarchical learning in AI-framework, and it has become the state-of-the-art method for various computer vision tasks. Furthermore, CNN applications are growing to encompass enhanced signal processing, specifically spatiotemporal signal recognition. Spatiotemporal data are related to both space and time, and it is displayed by data points of a 2D vector. CNN processes these data while preserving their spatiotemporal characteristics. Each spatiotemporal data point is processed throughout the CNN network by convolution layers followed by pooling layers and topped up with fully connected layers. A mathematical representation of a convolution operation in one dimension with an input vector $x$, a kernel $\omega$ with $i$, $d$ to denote iterators, and $(\circ)$ to denote the element-wise multiplication, can be expressed as $C(i)$ with $i$ is the index of an element in the new feature map:

$$C(i) = (\omega \circ x)[i] = \sum_d x(i-d)\,\omega(d) \qquad (1)$$

Since Spatiotemporal data is a 2D vector, the convolution operation in Eq. (1) can be extended to 2D. With a 2-D input $x$ and a 2-D kernel $\omega$ with $(i,j),(d,k)$ are iterators, the mathematical representation of a convolution in two dimensions can expressed as $C(i,j)$ with $(i,j)$ is the index of an element in the new feature map:

$$C(i,j) = (\omega \circ x)[i,j] = \sum_d \sum_k x(i-d, j-k)\omega(d,k) \qquad (2)$$

In supervised training, the procedure is launched by initializing the weights randomly to learn a function $h(x)$ (called hypothesis) to a desired output function $f(x)$ to map the input $x$ to a desired output $y$. The model is trained on a batch of data for a number of iterations and validated on a batch of data to identify a performance measure $P$ in each epoch. $P$ is optimized by reducing the prediction error of $h(x)$ in mapping the input $x$ to the output $y$, which is expressed by the cost function $J(\theta)$. Here J is the cost and θ is the parameters $\omega$, $b$, with other learnable parameters that can be adjusted during training with b as the bias term. The cost function is approximated by using forward propagation. This allows the calculation and storage of intermediate variables (including outputs) for a neural network in consecutive order from the input layer to the output layer. The optimization function is used to minimize the cost function with the weights $\omega$ of $\theta$. Therefore, the error in our hypothesis $h(x)$ for a given set of parameters θ is found using the generalized cost function. The cost function or error function are the sum of loss functions, while the loss functions are defined based on predicting a data point to a label with error. The widely used cost function for supervised classification problem is binary cross entropy. Essentially neural networks models are trained to learn to predict data $D = \{(x_1, y_1), (x_2, y_2), \dots (x_M, y_M)\}$ to binary labels $y = 0$ or 1, these labels can be categorical for multi-class (one-hot encoded label, such that label 3 in 4 classes problem would be [0 0 1 0]). In this paper we used categorical cross entropy because the intention detection is a multi-class problem. The binary cross entropy loss ($J_{bce}$) formula can be expressed as follows:

$$J_{bce} = -\frac{1}{M}\sum_{m=1}^{M}[y_m \log(h_\theta(x_m)) + (1 - y_m)\log(1 - h_\theta(x_m)). \quad (3)$$

To calculate the cost function of the binary cross entropy,

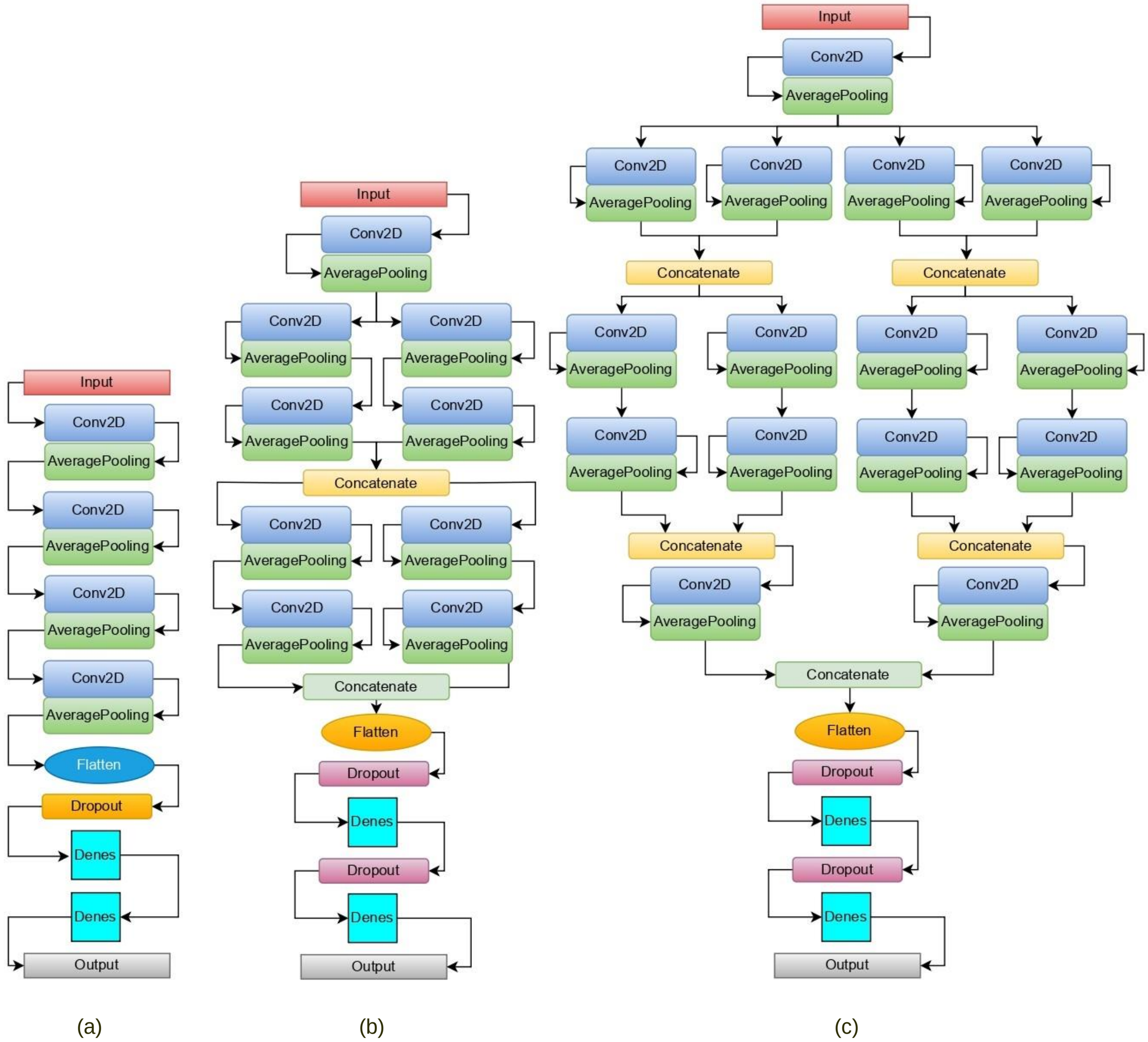

**FIGURE 7. Proposed CNNs networks architecture building block for spatiotemporal Kinova Gen3 robotic arm tasks classification. The color-coding of boxes: convolution layers and fully connected layer; pooling layers; concatenation layers and flattening layers; dropout layers; input at the top and softmax output layers. a) a single-stream CNN, b) a dual-stream CNN, c) a quadruplet stream CNN.**

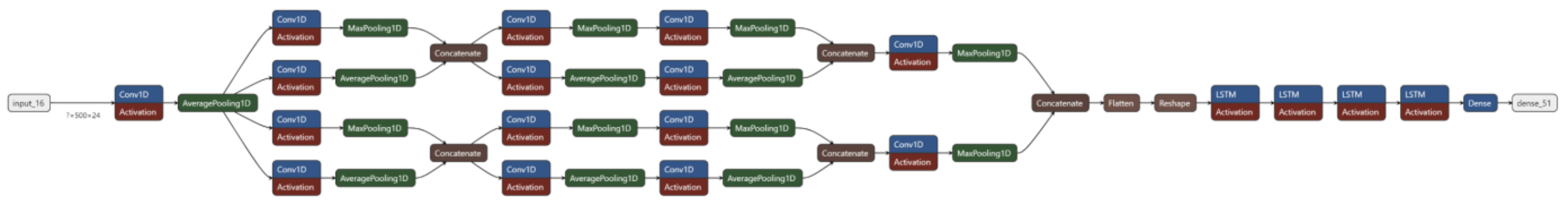

(a)

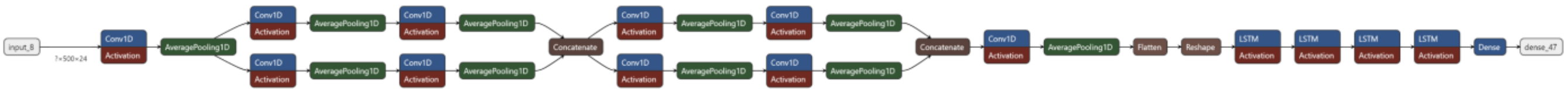

(b)

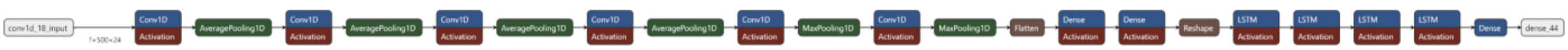

(c)

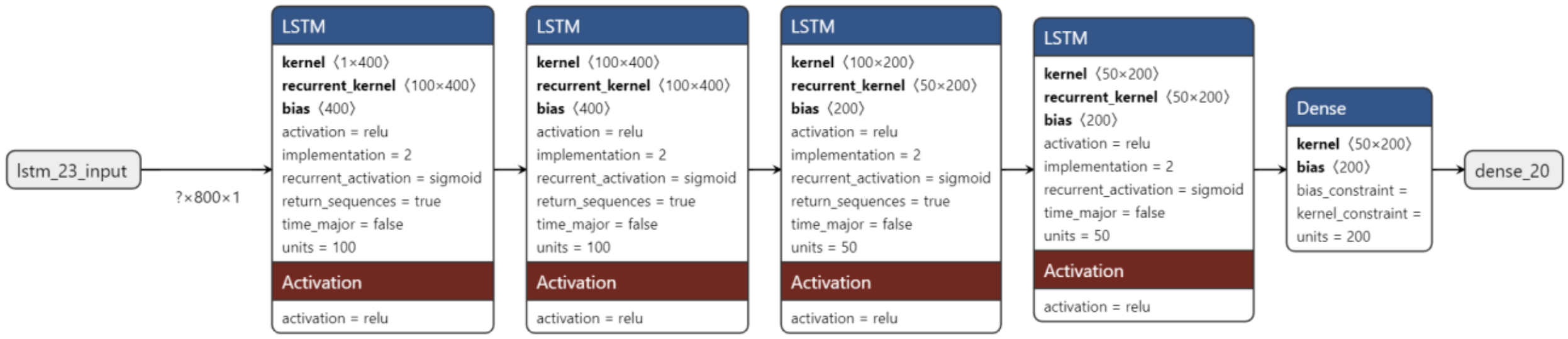

(d)

**FIGURE 8. Building block of the hybrid CNN+LSTM neural network for spatiotemporal data forecasting. a) Hybrid Quadruplet stream CNN+LSTM model, b) Hybrid dual-stream stream CNN+LSTM, c) Hybrid single-stream CNN+LSTM, d) LSTM architecture used in the CNN. The architectures are generated using the neural network visualizer '*Netron*' using the models' weights.**

we need to average the cross entropy across all $M$ data examples to make the loss function comparable across different size datasets. The categorical cross-entropy loss ($J_{cce}$), with $K$ to denote the number of classes and $y_m^k$the target label for training example $m$ for class $k$, is expressed as follows:

$$J_{cce} \;= -\frac{1}{M}\sum_{k=1}^{K}\sum_{m=1}^{M} y_m^k \, log \;(h_\theta \,(x_m \,, k)) \qquad (4)$$

### *F. LSTM model*

The Long Short-Term Memory (LSTM) model is utilized for forecasting robotic tasks, leveraging its capability to capture long-term dependencies in sequential data. By incorporating memory cells, LSTMs can retain information over time, effectively mitigating the vanishing gradient problem that often affects traditional recurrent neural networks. In our forecasting framework, the LSTM predicts future robotic tasks while continuously processing incoming robot force feedback. We segment the force feedback, using the first frames to forecast the next frames. This sequential prediction enables the LSTM to learn intricate patterns and relationships within the force feedback, facilitating accurate anticipation of future robotic actions. As the model engages with the force feedback, it iteratively updates its internal parameters based on the observed sequences, allowing it to grasp the underlying dynamics of robotic task performance. By analyzing historical patterns and trends in the force feedback, the LSTM can make informed predictions about upcoming tasks the robotic system should execute, thereby enhancing its ability to respond effectively to the current operational context.

In evaluating our LSTM model's accuracy, we employ the Root Mean Squared Error (RMSE) metric. RMSE measures the average magnitude of the errors between predicted and observed values, providing a comprehensive assessment of the model's performance. RMSE is calculated as:

$$RMES = \sqrt{\frac{1}{n}\,\sum_{i}^{n}(y_i - \hat{y}_i)^2} \qquad (5)$$

During the training phase, we optimize the LSTM model's parameters to minimize the RMSE. This optimization process involves iteratively adjusting the model's weights and biases using gradient descent-based algorithms, Adam is utilized in this study to improve its predictive capabilities.

### *G. Hybrid CNN+LSTM model*

The Hybrid CNN+LSTM model combines the strengths of Convolutional Neural Networks (CNNs) and Long Short-Term Memory (LSTM) networks to effectively forecast time series data. In this work, we propose a hybrid machine learning approach that combines Convolutional Neural Networks (CNN) and Long Short-Term Memory (LSTM) networks to detect the operator's intentions in robot-assisted glovebox telemanipulation. The model leverages spatiotemporal data recorded from the robotic arm's force feedback, effectively capturing the dynamics of various manipulation tasks. The CNN component plays a crucial role in extracting relevant features from the spatiotemporal data, identifying patterns associated with specific manipulation tasks through the analysis of the forces exerted by each joint over time. Following this feature extraction, LSTM layers are employed to analyze the temporal relationships within the data. LSTMs excel in handling time series data, retaining information over long sequences and enabling the model to understand how past actions influence future outcomes. By training on labeled data, the hybrid CNN-LSTM model learns to accurately categorize manipulation tasks. The operator's intentions are derived from the model's predictions, allowing for a nuanced understanding of the dynamics at play in telemanipulation systems. This integration of CNN and LSTM not only enhances the model's predictive capabilities but also improves its robustness in comprehending complex temporal dynamics, thereby advancing semi-autonomous manipulation in robotic systems.

## IV. EXPERIMENTS AND RESULTS

The data described in Section II.A. is preprocessed and utilized to train three different Convolutional Neural Network (CNN) models: the single-stream CNN, the dual-stream CNN, and the quadruplet-stream CNN, each of which serves a unique role in feature extraction (as depicted in Figure 7). The CNN networks consist of an input layer, convolution layers, pooling layers, fully connected layers, batch normalization, and an output layer with a softmax classifier. The set of stacked layers are Conv1D kernels (filter size × number of feature maps × number of filters), MaxPooling strides of 2 and pool size of 2. The models' classification performance is evaluated using confusion matrices and F1 scores for 20% of the data as detailed in the following experiments. These distinct CNN architectures are designed to capture varying levels of complexity and interaction within the data, ensuring that they learn nuanced features from the input. After the feature extraction process, the output from the final flattened layer of each CNN is fed into a Long Short-Term Memory (LSTM) network as shown in Figure 8 (a), (b), (c). The LSTM is specifically connected to the last flattening layer of the CNN, where it processes the extracted features to forecast temporal sequences within the data as shown in Figure 8 (d). The role of the LSTM layers is crucial, as they are trained to predict the temporal progression of the samples over time (the data are within time duration of 10 seconds), learning the underlying dynamics of the task being performed. With this approach, the LSTM analyzes the sequence of features to interpret the operator's intentions by processing an input sequence of steps and forecasting an output sequence of steps, learned from the robotic system's task manipulation. The input and output step sequences are then used to classify tasks based on data sets 2 and 3. Specifically, the input time frame data is combined with the forecasted output time frame data from the LSTM and fed into a multilayer perceptron (MLP) with a softmax layer for classification. Figure 6 illustrates the proposed framework of the CNN and LSTM and MLP.

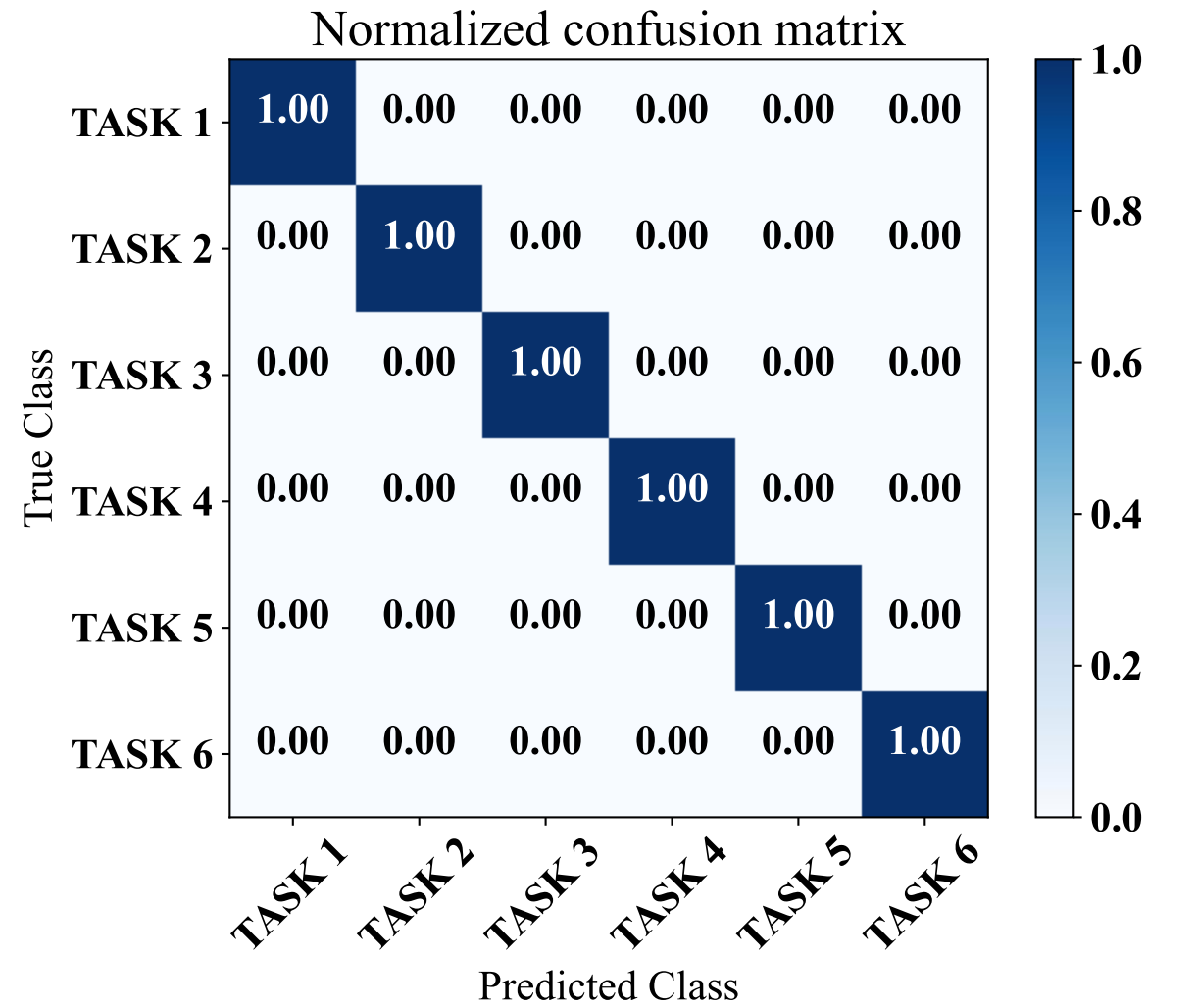

**FIGURE 9. Kinova-mic robotic arm classification confusion matrix dataset 1.**

TABLE. II
EXPERIMENTS IN FIG. 6 DESCRIPTION

| E | Description |
|---|---|
| E1 | Post object in glovebox & Radiation survey (dataset 2) |
| E2 | Place object in glovebox floor & Radiation survey (dataset 2) |
| E3 | Grasp radiation sensor & Radiation survey (dataset 2) |
| E4 | Grasp object and post out the glovebox & Radiation survey (dataset 2) |
| E5 | 6 tasks classes classification (dataset 2) |
| E6 | Place object in glovebox floor & Radiation survey & Grid radiation survey (dataset 2&3) |
| E7 | All tasks in dataset 2 as one class except radiation survey & Grid radiation survey in dataset 3 as second class |
| E8 | All tasks in dataset 2 as one class & Grid Radiation survey in dataset 3 as second class |

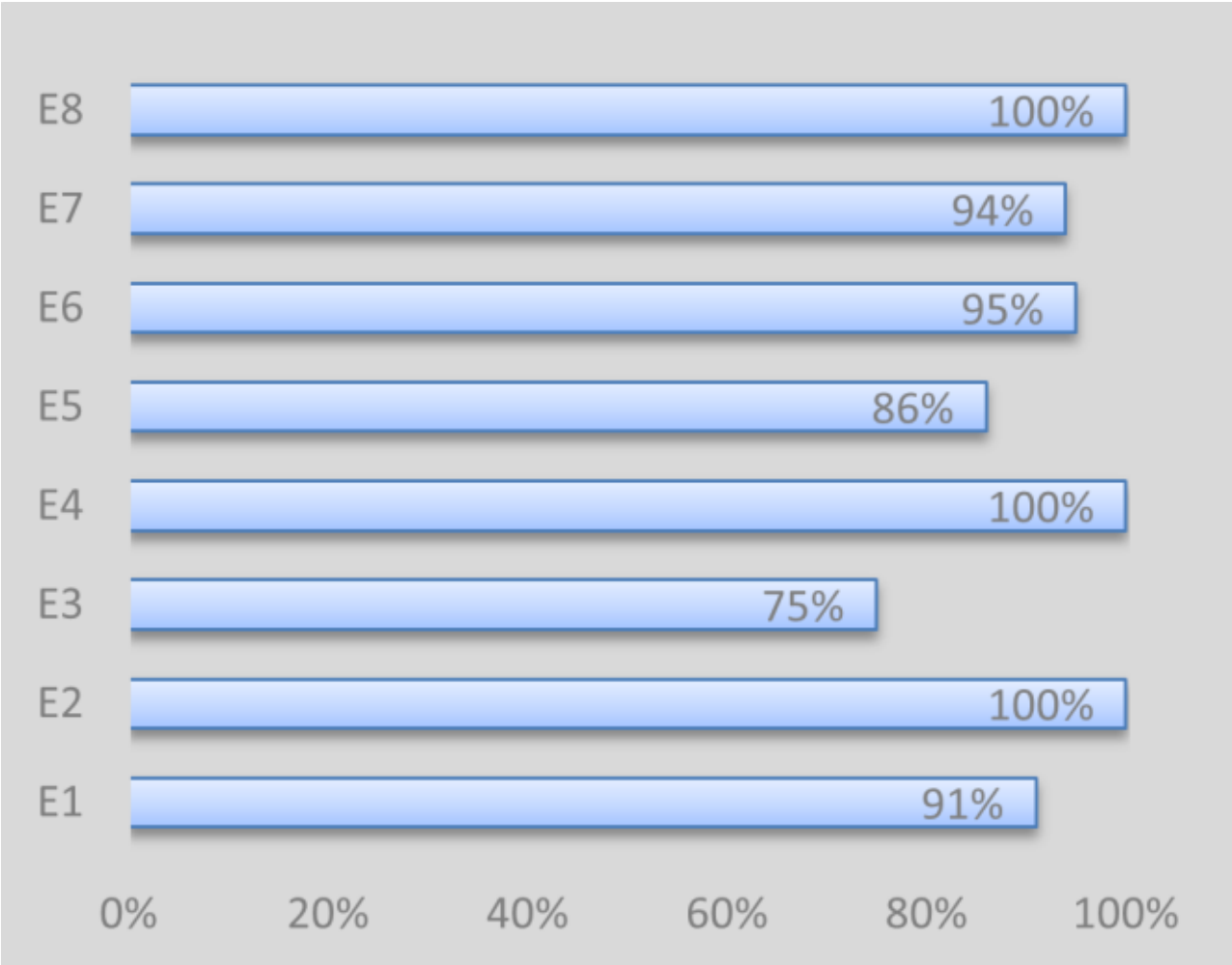

**FIGURE 10. Kinova robotic arm tasks receiver operating characteristic prediction.**

### *A. Data processing*

Accurate forecasting of robotic arm tasks is vital for effective intention detection. Therefore, the pre-processing pipeline was designed to ensure high-quality, standardized robotic arm force data analysis. The process involves converting the generated ROS bag files into CSV format, followed by further processing in Python to transform them into NumPy files. Each sample is meticulously labelled based on the manipulation task, resulting in the creation of a 1D vector of labels. Notably, the samples may vary in dimensions, particularly in the time domain (1D); thus, to ensure uniformity required by the CNN models, all samples are padded with zeros to achieve equal sizes. This preprocessing step is essential for effective model training and consistency. Consequently, the final NumPy combined dataset can be succinctly expressed as $x_{n,s} = [x_{n,1} \quad \cdots \quad x_{n,24}] \in \mathbb{R}^{n\times 24}$, where n is the number of the data block (1000 frames) and $s$ enumerates the spatial information, joints position, velocity, and forces. The recorded data measuring units varies, therefore, data standardization is implemented as a pre-processing step, to ensure that the data is internally consistent, such that the estimated activations, weights, and biases update similarly, rather than at different rates, during the training process and testing stage. The standardization involves rescaling the distribution of values with a zero mean unity standard deviation, using the following equation:

$$\widehat{x_{n,s}} = \frac{x_{n,s} - \mu(x_{n,s})}{\sigma(x_{n,s})} \tag{6}$$

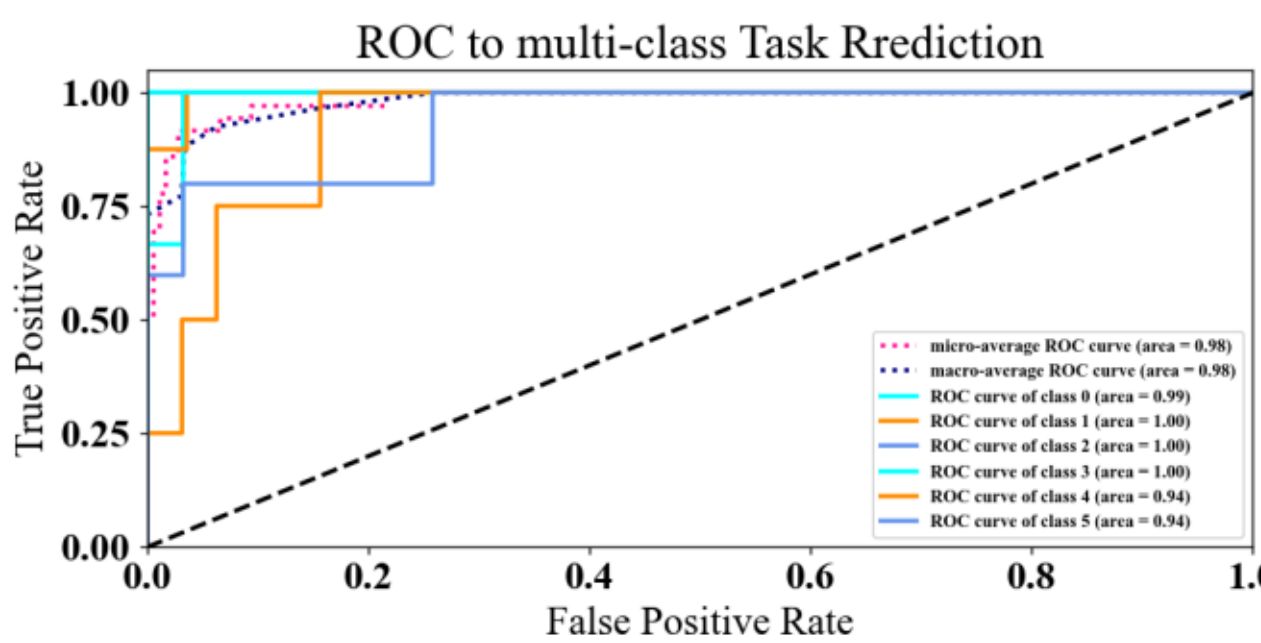

**FIGURE 11. Kinova robotic arm tasks receiver operating characteristic prediction, illustrating high prediction of 6 tasks.**

Here $\widehat{x_{n,s}}$ is robotic arm forces feedback data rescaled so that $\mu$ is the mean and $\sigma$ is the standard deviation. The CNN model (see Figure 7) is implemented to map the robotic arm's spatiotemporal data ($x(n,s)$) to an output label y by learning an approximation function $y = f\ (\widehat{x_{n,s}})$, $n$ denotes time and $s$ denotes data point recorded from the robot.

### *B. CNN Model for classification on Dataset 1*

The CNN model is trained and tested to detect the manipulation tasks detailed in Table 1, with 6 labels for a number of runs with random data split to ensure accurate classification. The first run was of the data 80% for training and 20% for validation and testing with 10% each. The second run was 70% for training and 30% for validation and testing with 15% each. The final run was 60% for training and 40% for validation and testing with 20% each. The CNN successfully classified the tasks in Table 1 by 100% - 2% F1-score, depending on data split for training and testing. The confusion matrix in Figure 9 shows the true positive predictions of the 6 classes for the first and second run, while the last run output was 98% F1 score.

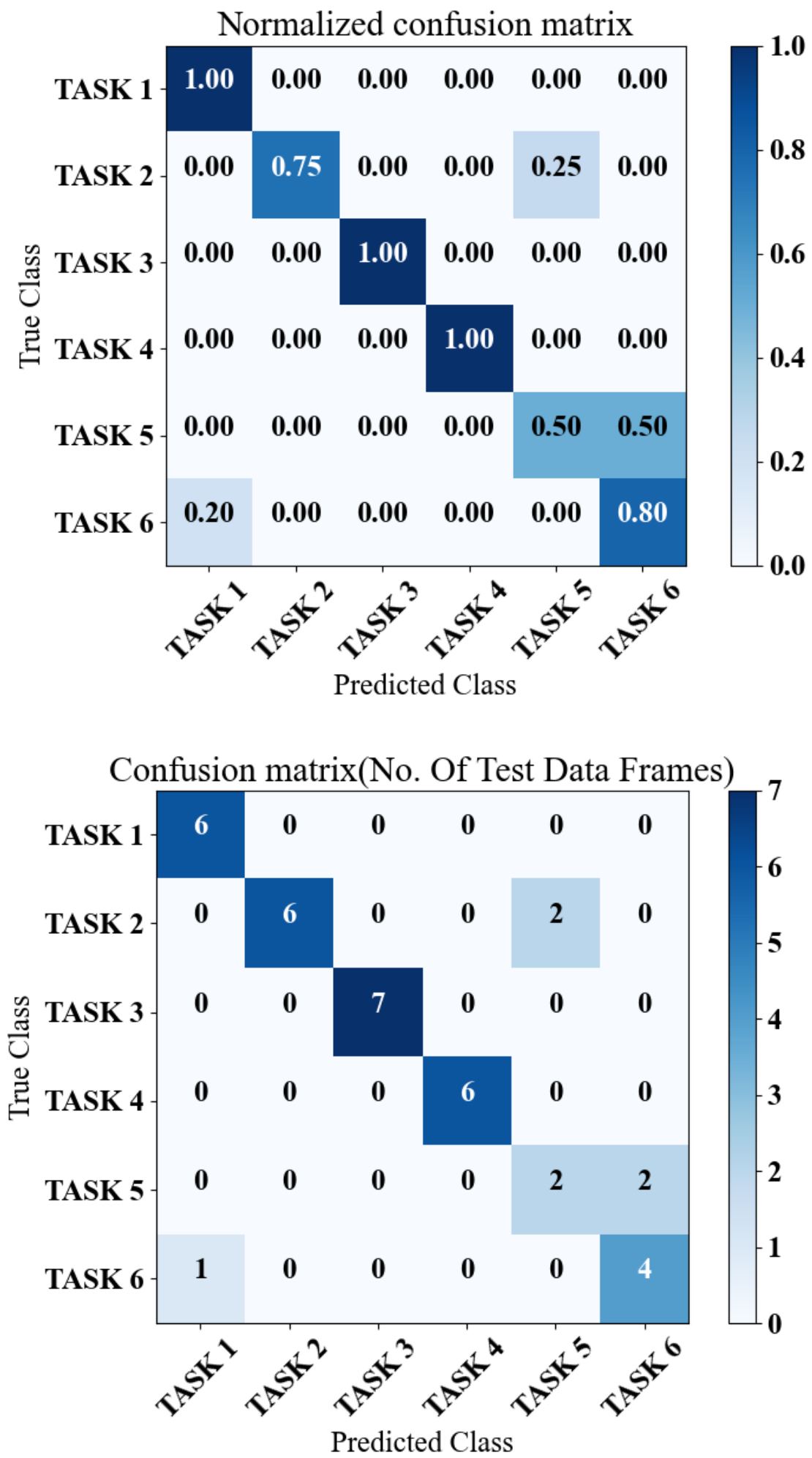

**FIGURE 12. Kinova robotic arm tasks classification confusion matrix dataset 2.**

### *C. CNN Model for classification on Dataset 2*

In the experimental setup, the model underwent a rigorous training and testing process, which spanned across two distinct folds. This approach ensured robustness and reliability in evaluating the model's performance. The primary objective was to equip the model with the ability to discern and categorize six manipulation tasks, each associated with specific labels meticulously detailed in Section II.A.1.

Through meticulous training and testing iterations, the model showcased promising results. Specifically, we trained three CNN models: the single-stream CNN achieved an impressive 86% accuracy, the dual-stream CNN attained 84% accuracy, and the quadruplet stream CNN excelled with 89% accuracy, as depicted in Figure 11 and Figure 12. These results were meticulously documented and visually presented in Table 2 and Figure 10, providing a comprehensive overview of the model's classification prowess.

A deeper exploration ensured, particularly in the discussion section, where the nuances of the model's performance were dissected and analyzed. Insights gleaned from this discussion shed light on the strengths and potential areas for improvement of the model, enriching the scientific discourse surrounding its efficacy. As the experimentation progressed, a pivotal refinement step was introduced in the form of a binary labeling system for the secondary dataset. This labeling scheme, thoughtfully designed, assigned a distinct label (0) to the task of grasping, while categorizing all other tasks under label (1). This strategic decision not only simplified the classification process but also facilitated a clearer delineation between the grasping task and its counterparts. With the implementation of this new labeling framework, the model underwent further training and testing. Specifically, it was trained and evaluated on tasks involving the placement of objects on the glovebox floor, radiation surveys, and grid radiation surveys using data sets 2 and 3. In Experiment E7, all tasks in data set 2 were treated as one class, except for the radiation survey and grid radiation survey in data set 3, which were considered as a second class. In Experiment E8, all tasks in data set 2 were treated as one class, while the grid radiation survey in data set 3 was treated as a second class. The outcomes were compelling, with the CNN model demonstrating exceptional classification performance. Across the majority of tasks, the model achieved an impressive F1-score of $100\% - 9\%$, underscoring its robustness and reliability in accurately classifying manipulation tasks.

TABLE III
LSTM & CNN+LSTM FORECASTING ON DATASET 2&3 MSE

| Model | Step In | Step Out | RMSE |
|---|---|---|---|
| LSTM | 500 | 500 | 0.065 |
| Single CNN+LSTM | 500 | 500 | 0.00991 |
| Dual CNN+LSTM | 500 | 500 | 0.0017 |
| Quadruplet CNN+LSTM | 500 | 500 | 1.91e-05 |

### *D. CNN Model for classification on Dataset 2&3*

In this experiment we fused dataset 2 with 3 in such a grid radiation Survey recorded from 6 operators is added to the binary labels in dataset 2. The CNN is trained and tested to classify object manipulation in the glovebox environment using these fused data. As detailed in Table 2 and shown in Figure 10, the model predicted the operator intention by 100% ± 6% F1 when object grasping compared to radiation survey and Grid radiation survey, demonstrated in the confusion matrix in Figure 10.

Nevertheless, in the pursuit of scientific rigor, it is imperative to acknowledge areas where the model exhibited room for improvement. Notably, in the case of grasp radiation sensor and radiation survey tasks, the model's F1-score dipped to 75%. While this discrepancy warrants further investigation, it also presents an opportunity for refinement and optimization in future iterations of the model.

Nevertheless, in the pursuit of scientific rigor, it's imperative to acknowledge areas where the model exhibited room for improvement. Notably, in the case of grasp radiation sensor and radiation survey tasks, the model's F1-score dipped to 75%. While this discrepancy warrants further investigation, it also presents an opportunity for refinement and optimization in future iterations of the model.

### *E. LSTM Model for forecasting on Dataset 2&3*

In this experiment, we analyzed temporal changes in task execution using the temporal frames (1000 frames representing 10 seconds) of the robotic arm's force feedback. An LSTM model was employed to interpret these temporal variations and forecast the tasks performed by the robotic arm over time. During training, the LSTM processed input sequences, predicted each task, and compared its predictions to actual values using the Root Mean Squared Error (RMSE) metric. Adam optimization was used to update the model's parameters based on Mean Squared Error (MSE), thereby enhancing its predictive accuracy. Model performance was evaluated on a validation set by calculating the MSE between predicted and actual values, providing insight into the model's generalization capability and reliability. The LSTM was trained on two datasets (Dataset 2 and Dataset 3) both datasets have 7 labels, with input data extracted from temporal frames, as detailed in the following sections.

#### *1) Spatial Mean LSTM*

The LSTM model (see Figure 8d) input was the dataset temporal frames in such for the data $x_{n,s} =$

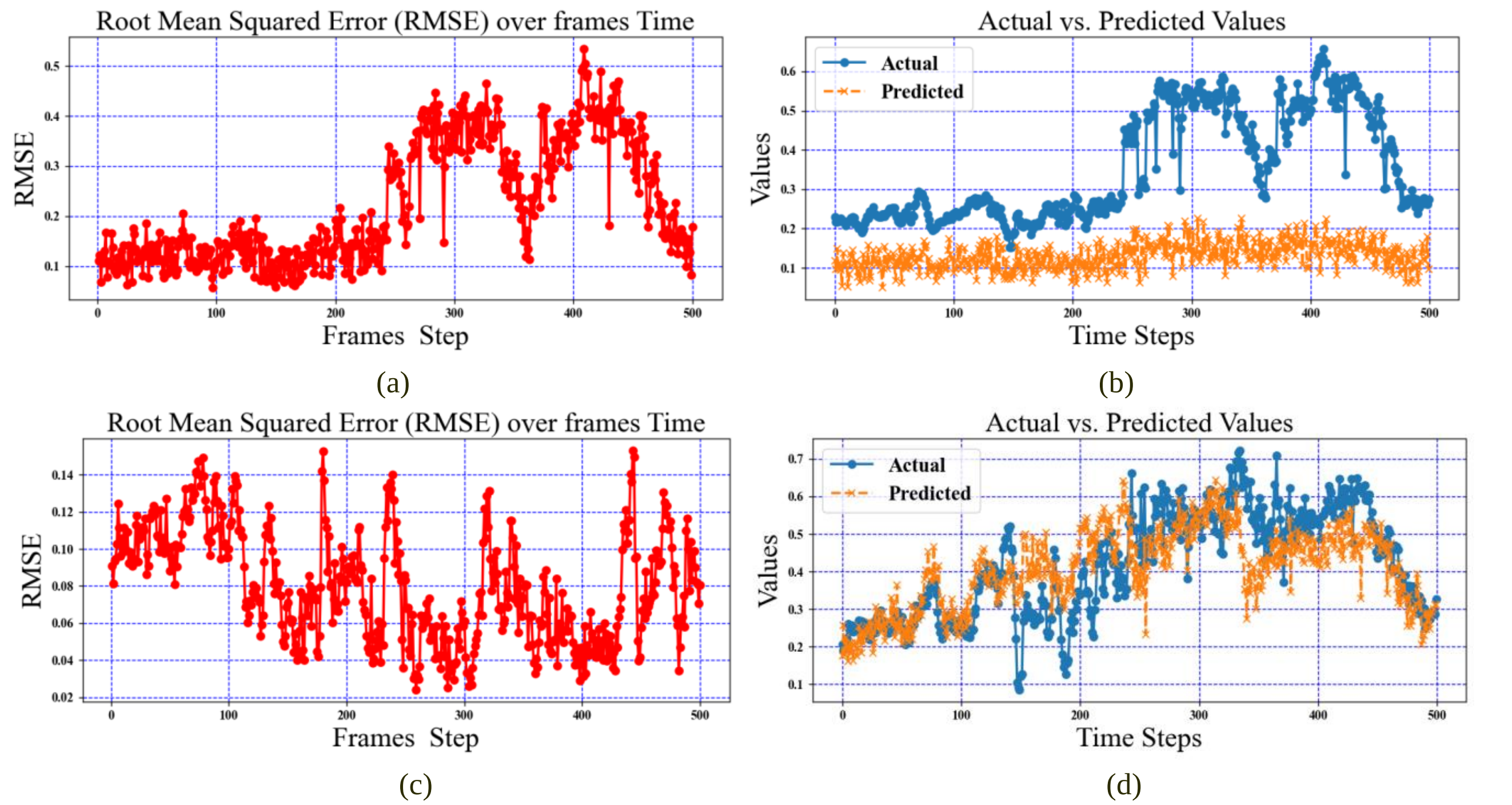

**FIGURE 13.** **LSTM model forecasting 500 frames for one sample. a) Root Mean Squared Error Over Time, b) Actual vs. Predicted values. With Mean Squared Error 0.06555 for all samples, 7 classes. Hybrid Single stream CNN+LSTM model forecasting 500 frames for one sample. c) Root Mean Squared Error Over Time, d) Actual vs. Predicted values. With Mean Squared Error 0.00991 for all samples, 7 classes**

$[x_{n,1} \quad \cdots \quad x_{n,24}] \in \mathbb{R}^{n\times 24}$, spatial domain is extracted using spatial mean which is calculated using the following formal:

$$SA[n] = \frac{1}{24}\sum_{i=1}^{24}(x_i[n]) \qquad (8)$$

Here $x_i$ represents the individual force readings from the robotic arm, and $n$ denotes the frames within each sample. The temporal domain was then used as input to the LSTM model to forecast the sequence of tasks. As shown in Table III, the forecasting was conducted with a step-in value of 500(5 second) and a step-out value of 500 (the remaining 5 second), resulting in a mean squared error (MSE) of 0.065. The performance of the temporal domain was suboptimal, which posed significant challenges for the LSTM in accurately forecasting the input. As illustrated in Figure 13(a), the maximum error value reached 0.5, while in Figure 13(b), the comparison between the predicted and actual outcomes indicates a misalignment.

*2) Hybrid Single CNN+LSTM*

The CNN ( see Figure 7 (a) is connected to the LSTM to form a Hybrid CNN+LSTM model, where the CNN acts as a feature extractor. Specifically, the CNN extracts the 1000 temporal frames from the dataset, denoted as $x_{n,s} = [x_{n,1} \quad \cdots \quad x_{n,24}] \in \mathbb{R}^{n\times 24}$ using convolutional and pooling layers, as shown in Figure 8(a). After training the CNN for classification (as detailed in Section IV.D), its parameters were frozen and used to fine-tune the CNN+LSTM, in such the final dense layers of the CNNs were replaced with LSTM layers for forecasting, while preserving the temporal domain. This approach enabled the hybrid model to effectively leverage the spatial features extracted by the CNN and capture temporal dependencies through the LSTM. The model's performance, evaluated using MSE, is presented in Table III. The results indicate that for a step-in of 500 (5 second) and step-out of 500 (5 second), the performance showed improvement compared to the spatial domain extraction presented in IV.E.1. However, the MSE remained relatively high, indicating only a slight enhancement in performance, as illustrated in Figure 13(c) (compared to Figure 13 (a), the RMSE is reduced). Figure 13(d) further demonstrates a marginal comparison between the predicted and forecasted sequences.

*3) Hybrid Dual CNN+LSTM*

Similar to the Hybrid Single CNN+LSTM, the Hybrid Dual CNN+LSTM was trained and evaluated in a comparable manner. The CNN depicted in Figure 7(b) was combined with the LSTM model, as illustrated in Figure 8(b), to forecast 1000 frames. The model's performance is presented in Table III. The results indicate that for a step-in of 500 and step-out of 500, the performance showed improvement compared to the *Dual CNN+LSTM* presented in IV.E.2. However, the MSE remained relatively high, indicating only a slight enhancement in performance, as illustrated in Figure 14(a). Figure 14(b) further demonstrates a marginal comparison between the predicted and forecasted sequences (compared to Figure 13 (d), the predicted is aligning with the actual in Figure 14 (b)).

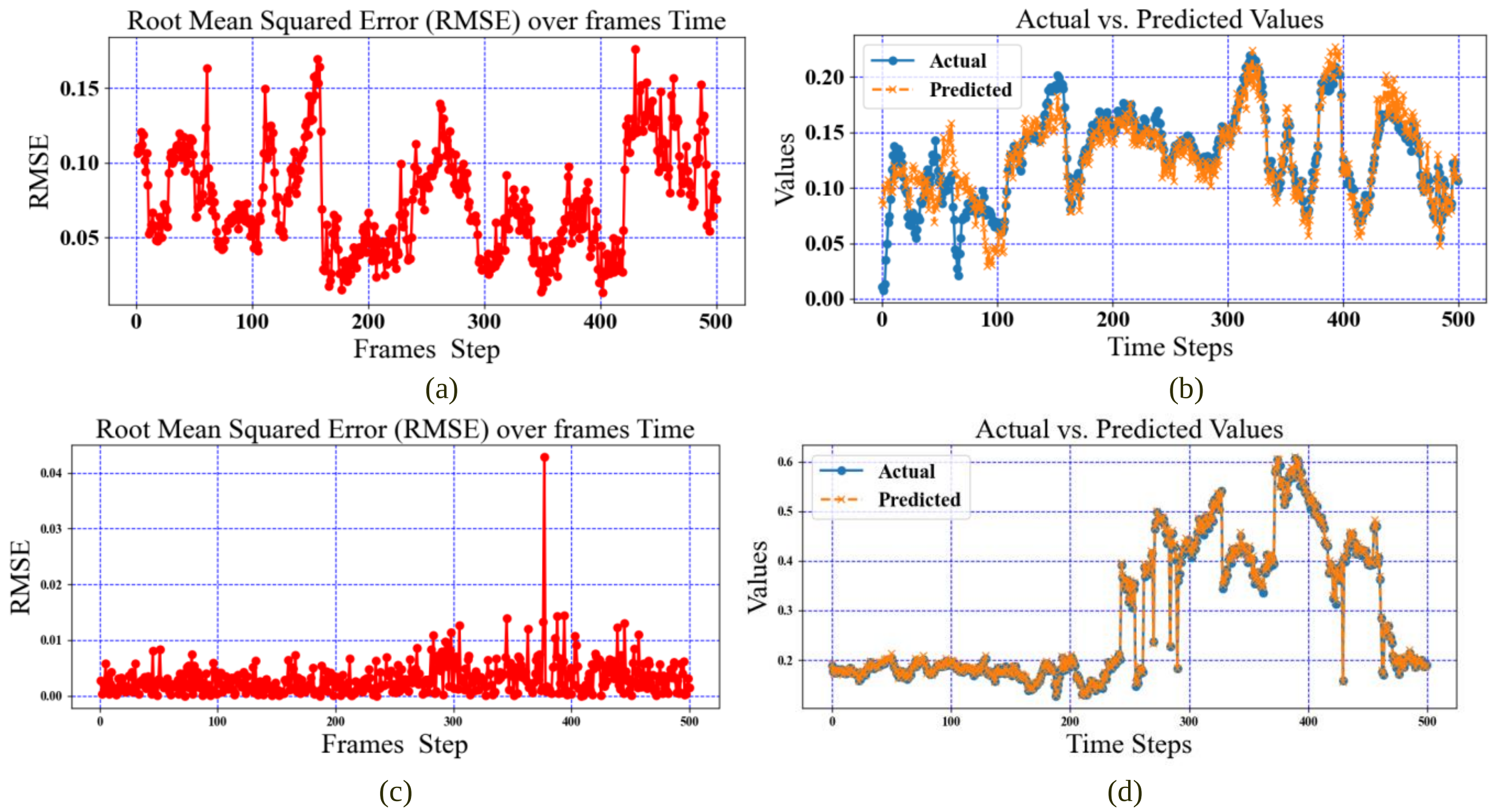

**FIGURE 14. Hybrid Quadruplet stream +CNNCNN+LSTM model forecasting 500 frames. a) Root Mean Squared Error Over Time, b) Actual vs. Predicted values. With Mean Squared Error 0.0017 for all samples, 7 classes. Hybrid Quadruplet stream CNN+LSTM model forecasting 500 frames. c) Root Mean Squared Error Over Time, d) Actual vs. Predicted values. With Mean Squared Error 1.9182e-05 for all samples, 7 classes.**

#### *4) Hybrid Quadruplet CNN+LSTM*

Similar to the Hybrid Single CNN+LSTM and Hybrid Dual CNN+LSTM, the Hybrid Quadruplet CNN+LSTM was trained and evaluated in the same manner. The CNN shown in Figure 7(c) was combined with the LSTM model, as depicted in Figure 8(c), to forecast 1000 frames, aiming to predict the robotic arm's movement in the glovebox. The Hybrid Quadruplet CNN+LSTM outperformed the previous hybrid models, achieving a lower RMSE. The best performance was obtained with Hybrid Quadruplet CNN+LSTM, resulting in an MSE of 1.91e-05, as shown in Table III. The comparison between predicted and actual values for the test datasets is reported in Figure 14 (c), (d). The peak RMSE value indicates that the error for a specific time frame was initially high. However, as shown in Figures 13 and 14, the RMSE values have now been reduced as shown in Figure 14 (c).

### *F. Forecasted Sequences Classifications*

The Hybrid Quadruplet CNN+LSTM model has demonstrated a notable ability to forecast the sequences of the robotic arm effectively. In this experiment, we classified the step-in and step-out by incorporating a multilayer perceptron (MLP) with a softmax layer at the output of the LSTM model. This approach involved combining the step-in with the predicted sequence and subsequently down sampling to classify the tasks based on categories E1 to E8, as presented in Table II and discussed in Section IV.D. The addition of the MLP is down sampled to a 1000 frames sequence to map it to the softmax layer robotic arm task probabilities. The results indicated a slightly lower F1 score, with a decrease of ± 0.07% F1-score, as evidenced by the confusion matrices shown in Figure 15, the F1 score for dataset 2 is 86%, where the original data was 89% F1-score.

## V. DISCUSSION

The presented study investigates the effectiveness of advanced modeling techniques in predicting operator intentions using raw spatiotemporal data from robotic arms. A convolutional neural network (CNN) is employed to automatically extract features related to manipulation tasks from data collected through a tele-manipulation system designed for a glovebox environment. We utilized CNN to classify various data from robotic arm manipulation tasks and performed a comparative analysis of classification performance between radiation surveys and object grasping alongside different autonomous manipulation tasks. Next, we compared the performance of various models for forecasting five seconds of data using a long short-term memory (LSTM) model. The output from the LSTM was employed for classification, enabling us to identify the operator's task while it was being executed. This capability is advantageous for anticipating the next task, allowing for the robotic arm to be positioned accordingly. As illustrated in Figure 1. (a), tools are arranged on different sides, and another robotic arm is present. The predictions from our model can engage this second robotic arm to prepare for the next task or be utilized to assess performance and detect hazards in the environment.

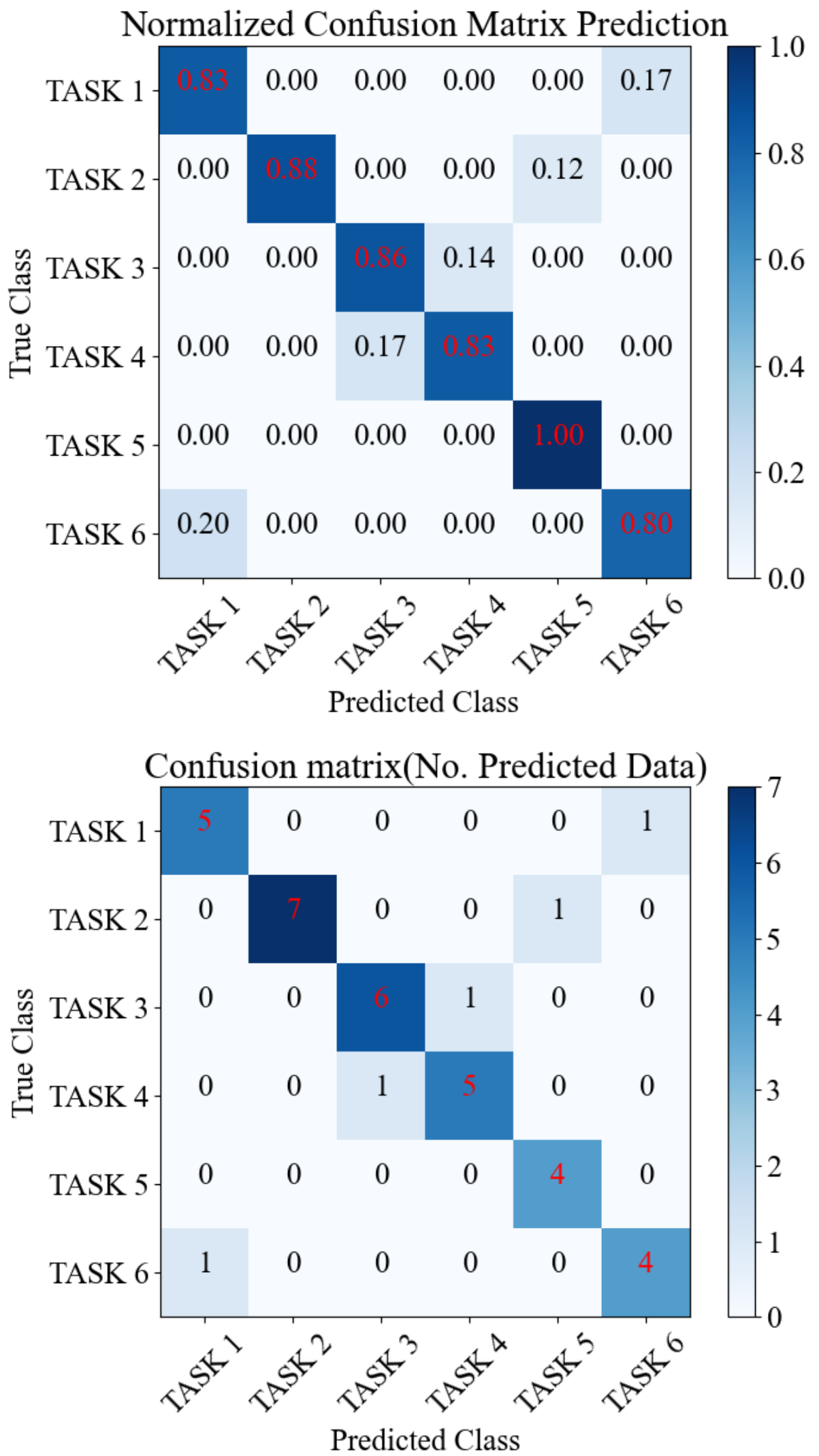

**FIGURE 15. Kinova robotic arm tasks classification from forecasted LSTM sequences confusion matrix for dataset 2 with 86% F1-score.**

To this end, a detailed discussion of the experimentation is presented in the following sections.

### *A. Autonomous Manipulation Classifications*

To test the validity of the proposed methods, we began our experimentation with the hypothesis that patterns can be learned from the robotic arm's movements and associated force feedback. The primary objective of employing data from autonomous manipulation tasks is to validate the accuracy of detecting manipulation patterns across multiple tasks based on raw spatiotemporal data. The model demonstrated a remarkable categorization capability, achieving an F1-score of 100% ± 2% across six distinct labels or classes, as illustrated in the confusion matrix (Figure 9). This high level of accuracy is attributed to the pre-planned joint positions and rotations, along with the corresponding force data signals, ensuring uniformity in spatiotemporal data characteristics across 20 trials for each task. Moreover, the ability of CNN to process complex spatiotemporal signals related to the robotic arm's movements and force feedback, as depicted in Figure 3, further motivated deeper investigation into this area.

### *B. Manual Manipulation Classifications*

The haptic interface (Virtuose 6D TAO) exhibits limited joint rotations, necessitating the disengagement of the interface between the haptic device and the robotic arm via a switch. This disengagement allows for offline adjustments to the haptic device's positions without affecting the Kinova arm's position. However, continuous adjustments to the haptic device can introduce variability in operator behavior, impacting the control of the robotic arm's end effector during task execution. The detection of operator intentions does not benefit from conventional methods, such as recording operator motion using inertial measurement unit (IMU) sensors, motion capture systems, or haptic device data. In this context, the recording of the Kinova arm's spatiotemporal data proves advantageous for accurately distinguishing specific manipulation scenarios and effectively detecting operator intentions.

Consequently, our research explores scenarios that are representative of real-world applications (see Figure 4), wherein objects are introduced into a glovebox environment for radiation surveys and subsequently removed upon completion of the inspection. This scenario, referred to as "dataset 2," involves a single operator utilizing a Kinova arm controlled by a haptic device. The identification of six distinct tasks by three CNN architectures not only enhances predictive capabilities but also highlights the nuanced performance across models. Notably, the single-stream CNN achieved an accuracy of 86%, while the dual-stream CNN demonstrated a commendable performance of 84%. However, the quadruplet stream CNN exhibited exceptional performance with an accuracy of 89%, as illustrated in Figure 10.

Despite the overall success, a detailed discussion regarding the challenges of task classification reveals complexities, particularly in scenarios characterized by minimal task variation, such as distinguishing between grasping tasks and the singular radiation survey task. This challenge is reflected in the moderate accuracy of 89% F1-scores observed in the six-class classification problem (see Figure 11 ROC curve and Figure 12 confusion matrix). While the models excelled in recognizing radiation survey tasks amid diverse grasping scenarios, the differentiation of subtle task nuances remains an area for improvement.

To address the variability in operator behavior stemming from continuous adjustments to the haptic device and to showcase the efficacy of the robotic arm, "dataset 3" was recorded with six operators engaged in real-world scenarios, specifically grid radiation surveys. The integration of raw spatiotemporal signals from "dataset 2" and "dataset 3" facilitated comprehensive training and testing of the CNN model, yielding accurate predictions of operator intentions, as detailed in Table 2 and Figure 10, experiments E6 to E8, with F1-scores of 94%, 95%, and 100%, respectively.

### *C. Manipulation Forecasting*

Subsequent experimentation focused on an LSTM model designed for forecasting operator intentions in "dataset 2" and "dataset 3." This model, tailored for predictive analysis, leverages force data derived from robotic arm manipulations to anticipate operator actions. By dynamically analyzing these signals, the LSTM model identifies patterns indicative of intended tasks, thereby enhancing real-time anticipation of operator intentions.

The evaluation of forecasting accuracy using the root mean square error (RMSE) metric (illustrated in Figures 13 and 14) underscores the LSTM model's capability to predict tasks by analyzing 500 frames of force data and forecasting the subsequent 500 frames. Lower RMSE values indicate superior alignment between predicted and observed tasks, showcasing the model's effectiveness in comprehending operator intentions and improving human-robot collaboration.

While our results demonstrate significant progress in task classification and operator intention recognition, it is essential to recognize inherent limitations and areas for future development. The moderate accuracy observed in distinguishing subtle task variations emphasizes the need for improved model robustness and adaptability to complex real-world scenarios. Additionally, further research aimed at refining the CNN and LSTM models to better encapsulate task nuances and operator intentions could lead to meaningful advancements in tele-manipulation systems and the overall effectiveness of human-robot collaboration.

### *D. Manipulation Forecasted Sequence Classification*

For real-time optimization in robotic tele-manipulation, it is crucial not only to predict the task an operator is currently engaged in but also to anticipate the next potential task. This foresight allows the system to adjust accordingly, improving both performance and efficiency. In glovebox environments, where operators work under high radiation levels, there is often the necessity for remote control operation rather than on-site, which introduces unique challenges, including issues such as signal delays. These delays can significantly impact the precision and fluidity of the manipulation process. To address these challenges, we incorporated task classification to optimize the control loop. By classifying tasks in real-time, delays can be mitigated through continuous semi-autonomous operation, allowing the robotic system to better respond to the operator's inputs despite environmental constraints.

In our experimental setup, we focused on classifying specific steps in the manipulation process, such as "step-in" and "step-out" sequence of the task. To achieve this, we used the first 5 seconds of recorded spatiotemporal data from the robotic arm as input to our model, and a long short-term memory (LSTM) network was employed to forecast the next 5 seconds. This forecast provides valuable predictive insights into the task flow, enabling the system to adjust in real-time. After the LSTM-generated output, we applied additional layers of a multilayer perceptron (MLP) to further refine the prediction of "step-in" and "step-out" phases.

The choice of using 5-second intervals was not arbitrary. Through experimentation, we found that most manipulation tasks in glovebox environments take between 8 to 10 seconds to complete. By analyzing various time intervals, we determined that 5 seconds was the optimal duration for input data, as it provided the best balance between accuracy and computational efficiency. This allowed the model to make timely predictions that closely matched the actual task execution process.

Once the tasks were classified, we compared the forecasted outputs with the actual recorded data to assess the accuracy of our predictions. The results were promising. Among the different models we evaluated, the quadruplet convolutional neural network (CNN) demonstrated the highest accuracy extracting features for the forecasting of tasks using the LSTM. As a result, we chose this model for our classification framework. Figure 15 illustrates the model's performance, where it achieved an F1-score of 86% when tested on Dataset 2. This high level of accuracy indicates that the model effectively predicted the operator's intentions and the corresponding robotic actions in real-time.

When we extended the analysis by combining Dataset 2 and dataset 3, we observed a slight decrease in classification performance, with the F1-score dropping by 2%. However, this reduction was still within acceptable limits, as the model maintained robust performance. For instance, as shown in Table II, Experiment E4 achieved an F1-score of 97%, while Experiment E8 maintained a perfect F1-score of 100%. These results demonstrate that the classification framework remains highly reliable across different datasets and tasks, even in complex and variable environments like gloveboxes.

By integrating task classification and prediction into the control loop, we have significantly improved the system's ability to handle delays and operator input variability. The LSTM-based forecast, combined with the additional MLP layers for task refinement, has allowed us to not only optimize current task execution but also to anticipate and prepare for subsequent tasks. This advancement is particularly useful in high-stakes environments, such as those involving radiation, where precision and timing are critical.

### *E. Comparison of This Work with Existing Research*

Uncertainty in robotic arm manipulation within gloveboxes arises from the robot's limited view of a constantly changing hazardous environment. Perceptual systems have the potential to reduce this uncertainty, enabling more reliable manipulation operations, making perception one of the key challenges in the field [25]. This study presents significant advancements in predicting operator intentions in tele-manipulation environments, surpassing traditional approaches like vision-based machine perception or tactile sensing [25]. The approach combines convolutional neural networks (CNN) for task classification with long short-term memory (LSTM) models for task forecasting, achieving high accuracy, with F1-scores reaching up to 100% in some instances. The quadruplet CNN architecture further enhances performance, particularly in challenging environments like high-radiation gloveboxes where signal delays and operational constraints are prevalent. Although the models performed exceptionally well in controlled settings, a slight decrease in performance was noted when datasets were merged, signaling the need for further exploration of their generalizability across varied real-world conditions. Nevertheless, the study's core contribution is the integration of CNN-based task classification with LSTM-based task forecasting, enabling real-time prediction of operator actions. This innovation supports smoother task execution, allowing robotic systems to adapt more effectively to operator inputs despite environmental limitations. Ultimately, by utilizing large databases containing examples of objects, material properties, tasks, and other relevant information [25], it enhances human-robot collaboration in hazardous environments, leading to safer and more efficient operations.

## VI. CONCLUSION AND FUTURE WORK

In this work, we propose a CNN and LSTM model to process raw spatiotemporal data from a robotic arm with the goal of accurately detecting the operator’s intention. Our model relies solely on the robotic arm's force feedback, which helps eliminate variability caused by environmental factors or the operator’s actions. By focusing on the arm's joint positions, velocities, and forces, the model predicts the operator’s intention through spatiotemporal data analysis. This targeted approach reduces the impact of external influences, enabling the system to generalize and optimize key features for other real-world manipulation tasks. Furthermore, the model can forecast and predict tasks during manipulation based on the force feedback, supporting real-time decision-making and control. Comparative analysis shows that the framework is highly effective in identifying manipulation tasks, with only a slight decrease of ±0.07% in F1-score compared to classification tasks, as shown in Table II and Figure 10. The model can be trained with diverse datasets, including those from different operators, robotic arms, and manipulation scenarios. This flexibility makes the approach suitable for various applications, such as manufacturing, healthcare, and space exploration.

There are several promising future directions, one potential avenue is collecting more diverse data from various operators and scenarios could enhance classifier performance and allow for better scalability. Future enhancements could include incorporating additional sensors on the robotic arm with coating to protect it from radiation to capture more precise manipulation tasks, enabling the system to better understand and interpret complex tasks. Additionally, to further enhance the model’s capabilities, future work will integrate the CNN+LSTM architecture into a reinforcement learning framework. This would allow the system to learn an optimal policy for semi-autonomous manipulation tasks. By using force data during manipulation, the model cannot only predict the next action but also autonomously execute subsequent manipulations, optimizing the entire task. This approach would enable the system to adapt dynamically during manipulation tasks, learning to perform actions autonomously based on feedback from the forces exerted.

## FUNDING STATEMENT

The authors extend their appreciation to the Deanship of Research and Graduate Studies at King Khalid University for funding this work through a Large Research Project under grant number RGP2/254/45. E. L. and G.H. thanks Project Robotics and AI Collaboration (RAICo) and Advanced Machinery & Productivity Initiative (AMPI). E. L., O.T., A.A. and G.H. thanks EPSRC Projects EP/R026084/1 Project RoL.,cs and Artificial Intelligence for Nuclear (RAIN), and EP/W001128/1 Robotics and Artificial Intelligence for Nuclear Plus (RAIN+).

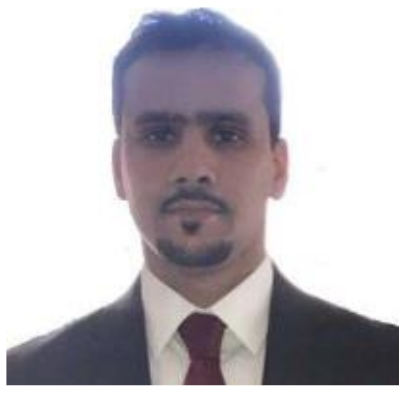

**ABDULLAH S. ALHARTHI** received his received the M.Sc. degree in electrical engineering from St. Mary's University, San Antonio, TX, USA, in 2015 and Ph.D. in Electrical and Electronic Engineering from The University of Manchester, U.K., in 2021. He currently serves as an Assistant Professor in the Department of Electrical Engineering at King Khalid University, Abha, Saudi Arabia. From 2015 to 2017, he worked as a Research Engineer in Robotics Technologies and Intelligent Systems at the King Abdulaziz City for Science and Technology (KACST) in Saudi Arabia. Following his Ph.D., he was a Research Associate in the Department of Electrical and Electronic Engineering at The University of Manchester from 2021 to 2022. He was postdoctoral fellow in the Electrical and Computer Engineering division at King Abdullah University of Science and Technology (KAUST) in Thuwal, Saudi Arabia, from 2022 to 2023. His research interests include pattern recognition, cognition, perception, theory of mind, gait analysis, and human-robot interaction.

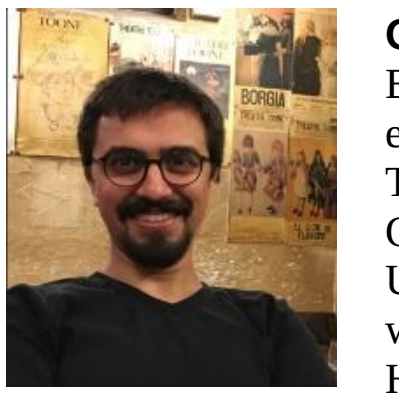

**OZAN TOKATLI** (Member, IEEE) received the B.Sc., M.Sc., and Ph.D. degrees in mechatronics engineering from Sabanci University, Istanbul, Turkey, in 2008, 2010, and 2015, respectively. Currently he is a Senior Robotics Researcher at the United Kingdom Atomic Energy Authority. He worked as a Postdoctoral Research Associate with Haptics Laboratory, University of Reading. He was a Research Engineer with RACE, UKAEA, which is a partner of the RAIN Hub. During his Ph.D., he worked on haptic systems with fractional order controllers and during his Postdoctoral research, his research focus revolves around the exciting field of physical human-robot interaction, where I delve into intricate control challenges related to haptics and teleoperation. At present, my work centers on the development of cutting-edge robot haptics technology, aimed at enhancing the resilience and agility of autonomous non-prehensile manipulation.

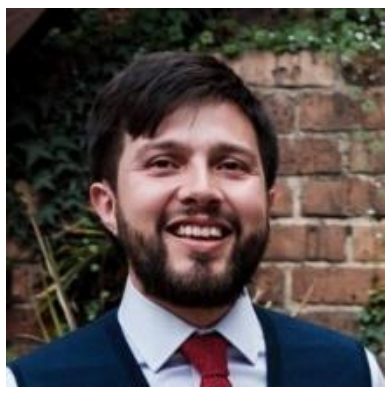

**ERWIN LOPEZ** (Member, IEEE) received the B.S. degree in Mechatronics Engineering from the Universidad Nacional de Colombia, Bogota, Colombia, in 2012 and his Ph.D. degree in Mechanical Engineering from University of Bristol, Bristol, UK, in 2019. He is currently a Postgraduate Research Associate at The University of Manchester, working on robotics and HRI for challenging environments. His current research interests include human robot interaction, data-driven modelling, teleoperation and autonomous systems.

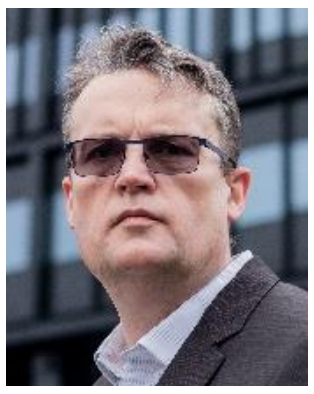

GUIDO HERRMANN (Senior Member, IEEE) received the degree "Diplom-Ingenieur der Elektrotechnik" (Highest Hons) from the Technische Universitaet zu Berlin, Berlin, Germany, in 1997 and the Ph.D. degree in control from the University of Leicester, U.K., in 2001. From 2001 to 2003, he was a Senior Research Fellow with the A*Star Data Storage Institute, Singapore. From 2003 to 2007, he was a Research Associate, Research Fellow, and Lecturer with the University of Leicester. Between 2007 and 2019, he worked at the University of Bristol, U.K., as a Lecturer, Senior Lecturer, and Reader in control and dynamics. Since 2019, he has been a Professor in Robotics and Intelligent Control at the University of Manchester. He is an Honorary Professor at the University of Bristol. His research interests include the development and application of novel, robust and nonlinear controllers with high-impact applications in robotics, automotive and high-precision systems. Dr. Herrmann is a Fellow of the IET and has been awarded with Dr. Jing Na the 2017 Hsue-shen Tsien Paper Award.